\documentclass[journal]{IEEEtran}

\usepackage{amssymb,amsmath,amsfonts,algorithm,algorithmic,subcaption,graphicx,cite,color,flushend,mysymbol,hyperref,mathrsfs,lipsum}
\graphicspath{{figures/}}
\begin{document}


\title{Model-Based Active Source Identification in Complex Environments}

\author{Reza~Khodayi-mehr,~\IEEEmembership{Student Member,~IEEE},  Wilkins~Aquino, and~Michael~M.~Zavlanos,~\IEEEmembership{Member,~IEEE}%
\thanks{Reza Khodayi-mehr, Wilkins Aquino, and Michael M. Zavlanos are with the Department of Mechanical Engineering and Materials Science, Duke University, Durham, NC 27708, USA, {\tt\footnotesize \{reza.khodayi.mehr, wilkins.aquino, michael.zavlanos\}@duke.edu}.
This work is supported in part by the NSF award CNS $\#1261828$.}
}

\markboth{}{Reza~Khodayi-mehr,~Wilkins~Aquino,~and~Michael~M.~Zavlanos} 

\maketitle

\begin{abstract}
We consider the problem of Active Source Identification (ASI) in steady-state Advection-Diffusion (AD) transport systems. Unlike existing bio-inspired heuristic methods, we propose a model-based method that employs the AD-PDE to capture the transport phenomenon. Specifically, we formulate the Source Identification (SI) problem as a PDE-constrained optimization problem in function spaces. To obtain a tractable solution, we reduce the dimension of the concentration field using Proper Orthogonal Decomposition and approximate the unknown source field using nonlinear basis functions, drastically decreasing the number of unknowns. Moreover, to collect the concentration measurements, we control a robot sensor through a sequence of waypoints that maximize the smallest eigenvalue of the Fisher Information matrix of the unknown source parameters. Specifically, after every new measurement, a SI problem is solved to obtain a source estimate that is used to determine the next waypoint. We show that our algorithm can efficiently identify sources in complex AD systems and non-convex domains, in simulation and experimentally. This is the first time that PDEs are used for robotic SI in practice.
\end{abstract}


\IEEEpeerreviewmaketitle

\section{Introduction} \label{sec:intro}
\IEEEPARstart{T}{he} problem of Source Identification (SI) refers to the estimation of the properties of a source using a set of measurements of a quantity that is generated under the action of that source. 
The SI problem has various applications ranging from environmental protection to human safety. Locating atmospheric, underground, or underwater pollutants, finding the source of a hazardous chemical leakage, and fire detection are a few examples.
In addition, SI can be an important component in higher level tasks such as search and rescue missions and crowd evacuation.

Due to its importance, the SI problem has been investigated extensively in the literature. Many approaches are heuristics that neglect the physics of the problem or model-based methods that are computationally expensive given the resources available to a mobile robot.
In this work, we propose an Active Source Identification (ASI) algorithm that, to the best of our knowledge, is the first method to systematically address the SI problem for a mobile robot utilizing the physics of the underlying transport phenomenon. 
Particularly, we focus on the steady-state Advection-Diffusion (AD) model as the running example, although, the same principals apply to any other transport model, e.g., heat transfer or acoustics.

\subsection{The Source Identification Problem} \label{sec:introSI}
%
The SI problem in AD transport systems is known as chemical plume tracing and odor localization in the robotic literature and has been investigated since the early $80$s. This task often entails three steps: detection, localization, and declaration, and most of the available algorithms focus on the localization stage \cite{ROL2008KR}. The algorithms differ depending on the dispersal mechanism, i.e., diffusion- or turbulence-dominated, and are specialized for the particular types of sensors used to take the measurements. They are often bio-inspired and try to mimic the behavior of different bacteria \cite{OMRN2002MND}, insects \cite{CRRCA2003RBSW}, or crabs \cite{BAASGTCP2012WVW}.
Generally, the main idea is to stay in the plume and move upwind, in the concentration ascent direction, or a combination of the two. In the literature, the former approach is called anemotaxis while the latter is called chemotaxis \cite{ROL2008KR}.

The authors in \cite{OMRN2002MND} propose a controller that combines anemotaxis and chemotaxis to localize a source. This method is compared to our algorithm in Section \ref{sec:sim}.
Arguing that gradient based methods can get trapped in local optima and plateaus, \cite{BREM2004DSR} proposes a biased random walk strategy for a robotic swarm to localize multiple point sources.
The authors in \cite{DRACPT2005ZSS} propose fluxotaxis which uses the mass conservation principle to trace a chemical plume using a robotic swarm.
In a more recent work \cite{DFOSLA2015SAPM}, a group of mobile agents are controlled to stay in a formation centered in the plume while they move upwind and localize an ethanol source. To localize multiple sources, the authors of \cite{VCPMOSL2013CMG} construct a statistical model of the discovered sources allowing the robots to find the next source by subtracting the effect of the previous ones.

When the transport phenomenon is turbulent, disconnected and non-smooth concentration patches appear in the medium. In this case, gradient-based approaches are often significantly inaccurate. The authors in \cite{infotaxis2007VVS} propose a gradient-less search strategy, called infotaxis, that maximizes the expected rate of local information gain. The authors in \cite{MSSLTM2016HHS} extend the infotaxis strategy to a multi-agent system. This approach shows a behavior that resembles that of a moth, i.e., casting and zigzagging, which amounts to an extensive exploration of the domain and can be energy-inefficient for a mobile sensor.

The above heuristic approaches to SI are often successful in practice but they also suffer from various drawbacks: First, they do not offer a systematic approach that can handle the localization task under a wide range of conditions. Instead, they are specialized for specific scenarios and sensors.
Second, these methods often can only localize a single point source or at best multiple point sources and provide no information about the intensity of the sources. Moreover, they declare localization when the robots physically reach the source while in fact it might be unsafe to approach the source in some applications.
Finally, often these heuristics are proposed for convex environments and they do not handle obstacles and non-convex domains easily.
%
These limitations can be addressed if the underlying physics is properly incorporated in the formulation, which leads to model-based SI methods. \footnote{Note that some of the heuristics above selectively utilize physical principles but our intention here is a dedicated, systematic formulation.}
These model-based SI methods are a special class of Inverse Problems (IPs) which have been studied for a long time; see, e.g., \cite{DIP2010H}. Methods to solve IPs rely on a mathematical model of the underlying transport phenomenon which often is a Partial Differential Equation (PDE) and, in the special case of SI problems, it is linear in the unknown source term.

The literature on model-based SI problems can be classified in different ways based on the state of the problem, the number of sources, and their shape.
Generally, transient transport phenomena are more challenging compared to the steady-state ones, but time-dependent measurements are more informative.
%
The localization of a single point source in steady-state is considered in \cite{SLSDENAD2005MGK} for a semi-infinite domain, whereas the authors in \cite{MBSTSLP2000AS} address the problem for transient transport relying on the \textit{a priori} knowledge of the possible point source locations.
%
SI in the presence of multiple point sources is considered using optimization-based methods.
For instance, the work in \cite{MSDLADPUWSN2009WSK} addresses the detection and localization of multiple such sources using a wireless sensor network.

More general problems that involve sources of arbitrary shapes in arbitrary domains are typically solved numerically using, e.g., the Finite Element Method (FEM).
Discretization of a steady-state PDE using the FEM, leads to a linear time-invariant system where the source term acts as a control input. From this perspective, the IP is similar to the problem of input reconstruction. However, one of the main assumptions in this problem is that the number of observations is no less than the dimension of the unknown input \cite{SEUIRRHSMO2012Z}. This assumption is violated in IPs which are typically ill-posed. To resolve this issue, regularization techniques can be used \cite{DIP2010H,meC4}.
The authors in \cite{VFEMSICDT2003ABGLW} use the FEM along with total variation regularization to solve the SI problem.
Similarly, in our previous work \cite{meC1}, we proposed the Reweighted Debiased $\ell_1$ algorithm, which is an iterative sparse recovery approach to the SI problem. Despite generality, numerical methods such as FEM become computationally demanding as the size of the domain grows.
Furthermore, unlike heuristic methods, none of the model-based SI methods discussed above, rely on robots to collect the measurements that are needed to solve the problem.

\subsection{Active Sensing} \label{sec:introOSP}
%
Optimal measurement collection has been long studied in the robotics literature to solve state estimation problems.
%
Given a probabilistic model of the measurement noise, information-theoretic indices, e.g., covariance \cite{meJ2}, Fisher Information Matrix (FIM) \cite{EEDI2016MSMM}, different notions of entropy \cite{PFBIAS2010RH}, mutual information \cite{MIPPASEP2014CLD,DASSSMRN2015ALP}, and information divergence \cite{SMPFT2011AL}, have been used for general robotic planning.
For example, given an information distribution, the authors in \cite{EEDI2016MSMM} propose an optimal controller to navigate the robot through an ergodic path. We investigate the performance of this planning method for SI in Section \ref{sec:sim}.
A common predicament in applying some of these methods for SI is the need for the posterior distribution of the unknown source parameters. Obtaining this distribution for SI problems requires solving stochastic IPs and is computationally expensive, see, e.g., \cite{SRM2010FMWB}. This makes the application of optimality indices that require calculation of the expected information gain, e.g., entropy, mutual information, and information divergence, intractable.

Typically in SI problems, the amount of information provided by a measurement depends on the value of the unknown parameters in addition to the measurement location. A common approach to address this point is to combine the path planning for optimal measurement collection with the solution of the SI problem in a feedback loop which leads to Active SI methods. To solve the planning problem, scalar measures of the FIM can be used, as for state estimation; see, e.g., \cite[Ch. 2]{OMMDPSI2004D}. 
The difference is that in SI the unknown parameters cannot be obtained in closed form by a filter update, but instead they are obtained by the solution of an IP that is much more difficult to solve.
%
Specifically, the work in \cite{ASIGRPE2005CR} presents trajectory planning for an autonomous robot, utilizing the trace of the FIM, to identify parameters of a transient Advection-Diffusion model under an instantaneous gas release in an infinite domain.
%
Similarly, \cite{OSMSNDLSCS2008PU, OMSPPEDPS2008TC} propose continuous-time optimal control methods that utilize the determinant of the FIM for trajectory planning for IPs with a few unknowns; \cite{OSMSNDLSCS2008PU} considers the SI problem in transient state under the assumption that the noisy measurements are taken continuously, while \cite{OMSPPEDPS2008TC} is an extension of \cite{OSMSNDLSCS2008PU} for general IPs.
In a different approach, the authors of \cite{DPSUE2009BLTT} propose an adaptive SI algorithm to localize a single point source emphasizing on path planning in unknown, possibly non-convex, environments.
Common in the above literature on ASI is that the proposed methods avoid the solution of complex SI problems by either assuming very simple mathematical models for the SI problem that can be efficiently solved, or by assuming that the solution of the SI problem is provided \textit{a priori} and the goal of planning is to collect measurements that are then examined to find a source term that matches those measurements.
Therefore, these methods do not apply to general SI problems and for this reason they have also not been demonstrated in practice. To the best of our knowledge, our work is the first ASI method that considers active measurement collection for realistic SI problems.

\subsection{Proposed Method}
%
In this paper, we consider the problem of Active Source Identification in Advection-Diffusion (AD) transport systems in steady-state. As most path planning methods for state estimation, we propose a method that combines SI and path planning in a feedback loop. The difference is that here the estimation problem is not solved by a closed form filter update, but instead it requires the solution of a complex PDE-constrained optimization problem.

Particularly, given a set of noisy measurements, we formulate the SI problem as a variational regularized least squares optimization problem subject to the AD-PDE.
To obtain a tractable solution to this problem, we employ Proper Orthogonal Decomposition \cite{OFROMPOD2007A} to approximate the concentration field using a set of optimal basis functions.
Moreover, we model the source term using nonlinear basis functions, which decreases the dimension of the parameter space significantly, although at the expense of introducing nonlinearity.
Using these parameterizations, we approximate the functional formulation of the SI problem with a low dimensional, nonlinear, constrained optimization problem, which we solve iteratively utilizing the gradient and Hessian information that we explicitly provide. To initialize this nonlinear optimization problem, we rely on the point-source Sensitivity Analysis of the SI objective function \cite{SSASLSSLS2013SA}.

Assuming a small number of measurements are available to initialize the identification process, we determine a sequence of waypoints from where a mobile robot sensor can acquire further measurements by formulating a path planning problem that maximizes the minimum eigenvalue of the FIM of the unknown source parameters with respect to the noisy concentration measurements.
%
The integrated algorithm, alternates between the solution of the SI and path planning problems. In particular, with every new measurement the solution of the SI problem produces a new source estimate, which is used in the path planning problem to determine a new location from where a new measurement should be taken, and the process repeats.
By appropriately decomposing the domain, we show that the proposed algorithm can identify multiple sources in complex AD systems that live in non-convex environments. 

\subsection{Contributions}
%
To the best of our knowledge, this is the first model-based ASI framework that has been successfully demonstrated in practice. As discussed in Section \ref{sec:introSI}, existing literature on ASI includes either model-free bio-inspired heuristic methods or model-based approaches that employ simplifying assumptions, e.g., point sources in infinite domains, to mitigate the complexity of solving real SI problems. Compared to the heuristic approaches, i.e., chemotaxis, amenotaxis, fluxotaxis, and infotaxis, our proposed model-based SI method combines all these bio-inspired behaviors systematically in a general identification framework. Specifically, the concentration readings are explicitly modeled in the least squares objective while the gradient information (chemotaxis), velocity information (anemotaxis), and the first principals (fluxotaxis) are rigorously encapsulated in the AD-PDE. Finally, the information content of the measurements (infotaxis) are incorporated in the solution of the planning problem. On the other hand, compared to model-based approaches that rely on oversimplified models, our method can solve more general and realistic problems. We have shown that our method outperforms existing approaches both in simulation and experimentally.

\begin{figure}[t!]
  \centering
    \includegraphics[width=0.5\textwidth]{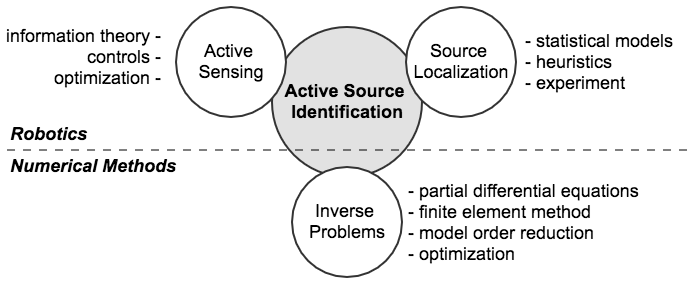}
        \caption{A schematic representation of the relevant literature.} \label{fig:literature}
\end{figure}

Like the path planning methods for state estimation, our method combines SI and path planning in a feedback loop. However, as discussed in Section \ref{sec:introOSP}, the difference is that unlike the estimation problem that is solved by a closed-form filter update, the SI problem requires the solution of a complex PDE-constrained optimization problem. The key ideas that enable a tractable solution to this problem are: (a) a suitable integration of model order reduction, point-source Sensitivity Analysis, and domain decomposition methods, (b) a nonlinear representation of the source term that reduces the dimension of the parameter space, and (c) an information theoretic  metric to measure the value of measurements for identifying unknown source parameters. The result is a set of techniques, insights, and methodological advancements that show how to efficiently design a model-based SI method that can be implemented onboard robots. Form a technical standpoint, the proposed framework bridges the gap in the rich but disconnected literature on source localization and active sensing that was discussed before; see Figure \ref{fig:literature}.

A significant contribution of this work is that it is the first to demonstrate applicability of robotic model-based SI methods to real-world problems. Real-world SI problems present major practical challenges related to modeling and estimation of the flow properties, which serve as the data in the AD-PDE, instability of the AD-PDE itself, and the uncertainties that are present in the parameters and boundary conditions; see the discussion in Section \ref{sec:discussion}.
We show that our algorithm is robust to uncertainties and performs well despite various simplifying assumptions made to model the real world; see \cite{TMCFD1993W,TD2002RW} for more details about these assumptions.

A preliminary version of this work can be found in the conference paper \cite{meC2}. The conference version only discusses the SI problem in convex domains and many theoretical details are absent due to space limitations; these details are included in Section \ref{sec:SI}.
In addition, here we discuss the integrated ASI algorithm that combines SI and planning and we also consider non-convex environments. Finally, we present experimental results that illustrate our method for real-world SI problems.

The rest of this paper is organized as follows. In Section \ref{sec:SI} we introduce the SI problem and discuss the proposed approach to solve it. Section \ref{sec:path} is devoted to the formulation and discussion of the path planning algorithm.
We discuss the integration of the SI and path planning algorithms along with their extension to non-convex domains in Section \ref{sec:autoSI}.
Section \ref{sec:sim} contains the numerical simulations and experimental results and finally Section \ref{sec:concl} concludes the paper.

\section{Source Identification} \label{sec:SI}
\subsection{Advection-Diffusion Transport} \label{sec:ADPDE}
Let $\Omega \subset \reals^d$ denote the domain of interest ($1 \leq d \leq 3$), and assume the presence of sources is modeled by a nonnegative function, $s: \Omega \rightarrow \reals_+$.
\footnote{For the problem considered here, we assume that sources are strictly positive functions. In general, sources can also be negative in the case of sinks. Sinks can appear, e.g., in the presence of chemical reactions that consume a contaminant. The treatment of the problem in that case is similar.}
Let $c: \Omega \rightarrow \reals_+$ be the measurable quantity, such as concentration, generated by this source function. Moreover, let the velocity at which this quantity is transported via advection be $\bbu \in \reals^d$ and $\kappa \in \reals_+$ be the diffusivity of the medium. Under steady-state assumption and applying a zero-valued Dirichlet condition to the boundaries $\Gamma$ of the domain, we arrive at the following Boundary Value Problem (BVP) \cite[ch. 2]{STMTPE2005SJ}
\begin{subequations} \label{eq:BVP}
\begin{eqnarray}
- \nabla \cdot ( \kappa \nabla c ) + \nabla \cdot (c \, \bbu) - s = 0 &\text{ in }& \Omega, \label{eq:ADPDE} \\
c = 0 &\text{ on }& \Gamma. \label{eq:EBC}
\end{eqnarray}
\end{subequations}
We consider Dirichlet conditions for the sake of simplicity; more general boundary conditions can be considered without any additional complications \cite{IFA1998R}.
In order for the BVP (\ref{eq:BVP}) to have a solution we assume that $s \in L^2(\Omega)$, i.e., $s$ is square integrable over $\Omega$, and define the feasible set for the source term as 
$ S = \set{ s \in L^2(\Omega) \ | \ s \geq 0 } $.

The BVP (\ref{eq:BVP}) can be equivalently represented in variational form as follows.
Consider the set $V \subset H^1_0(\Omega)$, i.e., the set of functions that themselves and their first weak derivatives are square integrable and have compact supports. Thus every $v \in V$ satisfies the boundary condition (\ref{eq:EBC}).
Multiplying equation (\ref{eq:ADPDE}) by the trial function $v \in V$, integrating over the domain, and using Green's theorem, we obtain the variational formulation of the Advection-Diffusion PDE as
\begin{equation} \label{eq:VBVP}
a(c,v) = \ell(v;s), \ \forall v \in V ,
\end{equation}
where $a: V \times V \to \reals$ is a non-symmetric continuous positive-definite bilinear form defined as
%
\begin{equation} \label{eq:binomial}
a(c,v) \triangleq \int_{\Omega} \kappa \nabla c \cdot \nabla v \ d\Omega +  \int_{\Omega} v \, \bbu \cdot \nabla c \ d\Omega ,
\end{equation}
\normalsize
and $\ell(s): V \to \reals$ is a continuous linear functional defined as
\begin{equation} \label{eq:functional}
\ell(v;s) \triangleq \inprod{\ell(s),v} \triangleq \int_{\Omega} sv \ d\Omega ,
\end{equation}
where the notation $\inprod{\ell(s),v}$ indicates the operation of $\ell(s)$ on the function $v$.
Given $s \in S$, we define the linear functional $\ccalM(c;s): V \to \reals$ as
\begin{equation} \label{eq:ADmodel}
\ccalM(c;s) \triangleq Ac - \ell(s) ,
\end{equation}
where the operator $A: V \to V'$ is defined by $\inprod{Ac,v} = a(c,v), \ \forall v \in V$. The notation $V'$ denotes the dual space of $V$, i.e., the space of linear functionals acting on $V$. Using this definition, the VBVP \eqref{eq:VBVP} is equivalent to the operator equation $\ccalM(c;s) = 0$ where $\ccalM: V \times S \to V'$.
Note that the functions $c$ and $v$ in the VBVP \eqref{eq:VBVP} have to be differentiable once.
Moreover, it can be shown that for $s \in S$ the BVP (\ref{eq:BVP}) and VBVP (\ref{eq:VBVP}) are equivalent and we can use them interchangeably. For further theoretical details, see \cite[ch. 8, 9]{IFA1998R}.

\subsection{The Source Identification Problem} \label{sec:probDef}
In this section, we formulate the SI problem as a constrained optimization problem subject to the AD transport model \eqref{eq:VBVP}.
Specifically, consider $m$ stationary sensors deployed in the domain $\Omega$ that take measurements of the concentration $c$, and let $E \subset \Omega$ be the set of $m$ compactly supported measurement areas enclosing the sensor locations.
\footnote{Note that the compact measurement area around any given sensor can be made arbitrarily small so that this sensing model approximates point measurements.}
Define, further, the indicator function $\chi_E: \Omega \to \set{0,1}$ for the set $E$ as
\begin{equation} \label{eq:indicFun}
\chi_E(\bbx) \triangleq 
\left\{
\begin{array}{ll}
1  &  \bbx \in E    \\
0  &  \bbx \notin E 
\end{array}
\right.
\end{equation}
and let $c^m: \Omega \rightarrow \reals_+$ be a function that assigns to $\bbx \in \Omega$ the noisy concentration measurement at that location, i.e.,
\begin{equation} \label{eq:measureFun}
c^m(\bbx) = \chi_E(\bbx) \, c(\bbx) \, ( 1 + \epsilon ), 
\end{equation}
where $\epsilon \sim {\cal{N}} (0, \ \sigma^2)$ and the measurement noise is proportional to the signal magnitude.
Then, the SI problem that we consider in this paper consists of determining an estimate $s$ of the true source term $\bbars$, given a set $E$ of $m$ noisy measurements in the domain $\Omega$, so that the AD model $\ccalM(c; s)=0$ defined in \eqref{eq:ADmodel} predicts the measurements $c^m$ as close as possible in the least squares sense.

The main challenges in solving the SI problem arise due to the following two reasons. First, generally the number of measurements $m$ is considerably smaller than the number of parameters that are used to describe the unknown source term. Second, the measurements are contaminated with noise.
To address these two challenges, we follow a standard approach and formulate the SI problem as a regularized least squares optimization problem subject to the AD model \eqref{eq:VBVP}. Let 
\begin{equation} \label{eq:Xnorm}
\norm{c - c^m}_{\chi_E}^2 \triangleq \int_{\Omega} (c - c^m)^2 \ \chi_E \ d\Omega 
\end{equation}
be a measure of discrepancy between the measurements and concentration field predicted by the AD model and define the cost functional $\ccalJ(c,s):V \times S \to \reals_+$ to be optimized by
$$ \ccalJ(c,s) \triangleq \frac{1}{2} \norm{c - c^m}_{\chi_E}^2 + \tau \ccalR(s) . $$
In this equation, $\tau$ is the regularization parameter and $\ccalR(s)$ is a functional that specifies the characteristics of the source $s$ that is selected as the solution of the SI problem.
In this work, we select
$ \ccalR(s) \triangleq \norm{s}_{L^1} = \int_{\Omega} \abs{s} \ d\Omega = \int_{\Omega} s \ d\Omega , $
where the last equality holds since $s$ is nonnegative. This choice of regularization penalizes the size of the source term.
Optimization of the objective functional $\ccalJ(c,s)$ subject to the AD model \eqref{eq:VBVP} gives rise to the following problem
\begin{equation} \label{eq:opt}
\min_{ (c,s) \in V \times S} \ccalJ(c,s) \ \, \st \ccalM(c,s) = 0,
\end{equation}
where the functional $\ccalM(c,s)$ is defined by \eqref{eq:ADmodel} and
\begin{equation} \label{eq:obj}
\ccalJ(c,s) = \frac{1}{2} \int_{\Omega} (c - c^m)^2 \ \chi_E \ d\Omega + \tau \int_{\Omega} s \ d\Omega .
\end{equation}

To solve the SI problem \eqref{eq:opt}, the gradient of the cost functional $\ccalJ(c,s)$ is needed. We obtain this gradient using the so called Adjoint Method. This method allows us to solve \eqref{eq:opt} directly in the reduced space $S$ of source functions rather than in the full space $V \times S$ of the concentration and source functions.
This is possible by using the model $\ccalM(c,s) = 0$ to represent the concentration $c$ as a function of the source term $s$, i.e., $c = \ccalF(s)$ where $\ccalF: S \to V$.
\footnote{As discussed in Section \ref{sec:ADPDE} such a representation exists and is unique.}
Using this gradient information we can minimize the cost functional $\bar{\ccalJ}(s) = \ccalJ(\ccalF(s),s)$ and determine the source term $s$ that solves the original problem \eqref{eq:opt}. See Appendix \ref{sec:adjointM} for the details of the Adjoint Method.

\subsection{Finite Dimensional Approximation} \label{sec:finiteDim}
The variables $c$ and $s$ of the optimization problem \eqref{eq:opt} are functions that live in the infinite dimensional function
spaces $V$ and $S$, respectively. Therefore, in order to solve this problem numerically, it is necessary to approximate $V$ and $S$ by finite dimensional subspaces $V_d  \subset V$ and $S_d \subset S$ determined by appropriate sets of basis functions. This approximation allows us to parametrize the concentration and source functions by a finite number of parameters that depend on the basis functions that constitute $V_d$ and $S_d$.

The key idea to obtain the finite dimensional subspace $V_d$ of the concentration function space $V$ is to use Proper Orthogonal Decomposition (POD) to reduce the order of the model.
The POD method is easy to implement and gives an optimal set of basis functions that can be readily used in our formulation to parameterize $c$. For a survey of popular model order reduction methods, see, e.g., \cite{ALDS2005A}.
At the same time, we use a nonlinear representation of the source term $s$ as a combination of compactly supported tower functions. This representation reduces the dimension of $S_d$ drastically, compared to classical approaches that utilize the Finite Element method.

\subsubsection{Model Order Reduction} \label{sec:MOR}
To reduce the order of a model using POD we need to solve the AD-PDE \eqref{eq:BVP} for all values of the unknown source term and build a set of basis functions that span the solution of the AD model.
We refer to the solutions as the snapshots of the problem.
Let $C = \set{c_i(\bbx)}_{i=1}^R$ denote a set of $R$ snapshots obtained by solving the AD-PDE \eqref{eq:BVP} for different realizations of the source term, i.e., each $c_i(\bbx) \in V$ corresponds to a given $s_i(\bbx) \in S$.
The objective of POD is to generate a set of optimal basis functions that maximize the averaged projection of the snapshots over these basis functions; see, e.g., \cite{PODRBFCPE2001AK}.
This optimization problem is equivalent to an eigenvalue problem for the covariance matrix $\bbC \in \reals^{R \times R}$ defined by
\begin{equation} \label{eq:PODcov}
\bbC_{ij} \triangleq \frac{1}{R} \int_{\Omega} c_i \, c_j \, d\Omega .
\end{equation}
The details of this procedure are presented in Algorithm \ref{alg:POD}, which yields $V_d = \text{span} \set{\psi_k}_{k=1}^N $ for $N < R$ where $\psi_k$ are the POD basis functions.
\begin{algorithm}[t]
\caption{Proper Orthogonal Decomposition}
\label{alg:POD}
\begin{algorithmic}[1]

\REQUIRE The set of snapshots $C = \set{c_i(\bbx)}_{i=1}^R$;

\STATE Construct the covariance matrix $\bbC$ using equation (\ref{eq:PODcov});

\STATE Solve the eigenvalue problem $\bbC \bbQ = \bbLambda \bbQ$ such that 		\label{line:Lambda}
$$\lambda_1 \geq \lambda_2 \geq \dots \geq \lambda_R \geq 0 \text{ and } \bbQ = [\bbq^1 \ \bbq^2 \ \dots \ \bbq^R] ; $$

\STATE The POD bases $\set{\psi_k}_{k=1}^R$ are given by 
\begin{equation}
\psi_k = \sum_{i=1}^R q^k_i c_i . 
\end{equation}

\STATE For $N < R$ the reduced order model $c_d$ is given as $c_d \in V_d = \text{span} \set{\psi_k}_{k=1}^N$.

\end{algorithmic}
\end{algorithm}
In line \ref{line:Lambda} of this algorithm $\bbLambda$ is the diagonal matrix of the eigenvalues.

As shown in \cite[thm. 1]{PODRBFCPE2001AK}, the $i$-th eigenvalue $\lambda_i$ of matrix $\bbC$ contains the average energy in the $i$-th mode. Moreover for a given number $N < R$ of basis functions, the POD bases have the maximum possible energy and are optimal.
Thus, for a given fraction $\eta$, we can select the number $N$ of required bases as the smallest number such that
\begin{equation} \label{eq:selecN}
\frac{\sum_{i=1}^N \lambda_i}{\sum_{i=1}^R \lambda_i} \geq \eta.
\end{equation}

\subsubsection{Parameterization} \label{sec:param}
Using the basis functions $\psi_k$ that constitute $V_d = \set{\psi_k}_{k=1}^{N}$ we can represent the functions $c$ and $v$ by a finite number of parameters, that can be used for numerical optimization. Specifically, we define
\begin{equation} \label{eq:finiteDim}
c_d = \bbpsi \,  \text{ and } \, v_d = \bbpsi \, \bbv ,
\end{equation}
where $\bbpsi = [\psi_1 \ \dots \ \psi_N]$ and $\bbc, \bbv \in \reals^N$.

To parametrize the source function $s$ we follow a different approach. Specifically, we propose a nonlinear representation of this term as a combination of compactly supported tower functions. The motivation for this representation is that each compactly supported source area can be approximately described by a very small number of parameters corresponding to the intensity and shape of the source. In this paper we focus on rectangular sources, although other geometric shapes can also be used for this purpose. 

\begin{figure}[t!]
  \centering
    \includegraphics[width=0.35\textwidth]{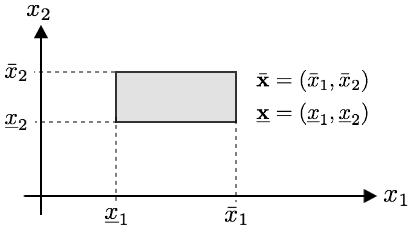}
        \caption{The support of a tower function in $2$D case defined by equation \eqref{eq:towerF}. } \label{fig:towerF}
\end{figure}

In particular, let $M$ be the number of basis functions used to approximate the source term in the domain $\Omega \subset \reals^d$ and consider two parameters $\set{ \undbbx_j, \barbx_j }$ for each basis function, where $\undbbx_j, \barbx_j \in \reals^d$ and $j \in \set{ 1, \dots, M}$. We define the compactly supported tower functions as
\begin{equation} \label{eq:towerF}
\phi_j(\bbx; \undbbx_j, \barbx_j) \triangleq
\left\{
\begin{array}{ccl}
  1    &   & \text{if }   \undbbx_j \leq \bbx \leq \barbx_j  \\
  0    &   &   \text{o.w.}
\end{array}
\right.
\end{equation}
where the inequalities are considered component-wise and $\undbbx_j \leq \barbx_j$; cf. Figure \ref{fig:towerF}. Then, for practical purposes we can approximate the desired source term $s \in S$ by
\begin{equation} \label{eq:sourceFiniteDim}
s_d(\bbx) = \sum_{j=1}^M \beta_j \phi_j(\bbx; \undbbx_j, \barbx_j) ,
\end{equation}
where we require $\beta_j \geq 0$ so that $s_d \in S$.
We denote by $ \bbp = (\beta_1, \undbbx_1, \barbx_1, \dots, \beta_M, \undbbx_M, \barbx_M)$ the vector of parameters associated with the source term $s_d$. Thus for $\Omega \in \reals^d$, $\bbp \in \reals^p$ where $p = M(2d+1)$.

Substituting the approximations $\set{c_d,s_d}$ of the concentration and source terms $c$ and $s$ in equation \eqref{eq:opt}, we obtain a finite dimensional counterpart of the SI problem as
\begin{align} \label{eq:optFD}
& \min_{ \bbc, \bbp} J(\bbc, \bbp) \\
& \ \st \bbM(\bbc, \bbp) = \bb0, \nonumber \\
& \ \ \ \ \ \ \beta_j \geq0, \ \bbl \leq \undbbx_j \leq \barbx_j \leq \bbu , \nonumber
\end{align}
where $j \in \set{ 1, \dots, M}$ and $\bbl, \bbu \in \reals^d$ are the lower and upper bounds on the coordinates of the domain.
\footnote{We can include additional convex constraints to confine the feasible region that might contain the true source. The extension to non-convex domains is considered in Section \ref{sec:autoSI}.}
Moreover, the objective $J:\reals^{N \times p} \to \reals_+$ is defined as $J(\bbc, \bbp) = \ccalJ(c_d, s_d)$ and the finite dimensional model $\bbM:\reals^{N \times p} \to \reals^N$ is defined as $\bbM(\bbc; \bbp) = \ccalM(c_d; s_d)$.

The optimization problem \eqref{eq:optFD} can be solved by a variety of available nonlinear optimization algorithms. Any such algorithm requires the first and possibly second order information, i.e., the gradient and Hessian of the objective function, as well as a proper initialization since the problem is nonlinear. 
In Appendix \ref{sec:SInumeric}, we derive explicit expressions for the gradient and Hessian of the objective function in reduced space $S_d$. Consequently, only the bound constraints in \eqref{eq:optFD} need to be considered explicitly for numerical optimization. In the same appendix, we also discuss the Sensitivity Analysis (SA) method for the initialization of problem \eqref{eq:optFD}.

\section{Mobile Robot Path Planning} \label{sec:path}
In Section \ref{sec:SI} we developed a way to efficiently solve the SI problem provided a set of stationary measurements is available.
In this section we propose a method to plan the path of a robot so that it collects the required measurements in a way that is optimal with respect to a desired optimality measure.
Specifically, we employ the Fisher Information Matrix (FIM).
Since the concentration field depends nonlinearly on the source parameters $\bbp$ via the AD model \eqref{eq:Model}, derived in Appendix \ref{sec:Grad}, the FIM depends on the unknown source parameter. Thus, selecting an optimal set of measurements requires feedback from the SI problem \eqref{eq:optFD} and couples the SI and planning problems.

The path planning problem is initialized with an initial set of $\bbarm$ measurements covering the domain of interest, that are necessary to obtain an initial estimate of the source locations and calculate the initial value of the FIM.
These measurements can be obtained by a stationary sensor network that can detect the presence or not of a concentration by monitoring the domain of interest. Optimal selection of these measurement locations is beyond the scope of this paper and here we assume that they are given; see, e.g., \cite{NSPMIMCC2006KGGK}.
After initialization, the robot collects new measurements sequentially so that the minimum eigenvalue of the FIM is maximized, providing in this way worst-case performance guarantees.

Let $\tbx_m = (\bbx^1, \dots, \bbx^m)$ for $m>\bbarm$ denote the sequence of measurement locations that belong to the set $E$, introduced in Section \ref{sec:probDef}, and define by $\bby_m \in \reals^m$ the associated vector of measurements, where from equation \eqref{eq:measureFun} we have $y_i = c^m(\bbx^i)$ for $i \in \set{1, \dots, m}$.
Approximating the concentration function $c$ with its finite-dimensional counterpart $c_d$ from Section \ref{sec:param}, we get
$$ y_i \approx c_d(\bbx^i; \bbp) (1 + \epsilon) , $$
where $\epsilon \sim \ccalN(0, \sigma^2)$ and we include $\bbp$ to emphasize that the concentration and consequently the measurements depend on the value of the source parameters. We can equivalently represent this equation as
$$ y_i \approx c_d(\bbx^i; \bbp) + \bbarepsilon(\bbx^i) , $$
where $\bbarepsilon(\bbx^i) \sim \ccalN (0, \bbarsigma^2 )$ and $\bbarsigma(\bbx^i) = \sigma c_d(\bbx^i)$.
In order to derive a closed form representation of the FIM, we make the conservative assumption that $\bbarepsilon(\bbx^i) \sim \ccalN (0, \bbarsigma_{\max}^2 )$ where $\bbarsigma_{\max} = \max_{\bbx \in \Omega} \bbarsigma(\bbx)$.
Then, the additive noise $\bbarepsilon(\bbx^i)$ is Normal, spatially independent, and identically distributed and the FIM is given explicitly as \cite{OMMDPSI2004D}
\begin{equation} \label{eq:defFIM}
\bbF = \frac{1}{\bbarsigma_{\max}^2} \sum_{i=1}^m \left( \frac{\partial c_d(\bbx^i; \bbp)}{\partial \bbp} \right)^T \left( \frac{\partial c_d(\bbx^i; \bbp)}{\partial \bbp} \right) , 
\end{equation}
where $\bbF \in \mbS_+^p$, i.e., $\bbF$ belongs to the space of $p \times p$ symmetric positive-definite matrices and $p = M(2d+1)$ is the number of unknown parameters.
The FIM provides a measure of the amount of information that the measurement vector $\bby$ contains about the source parameters $\bbp$. Note that the information values for independent observations are additive.

Recall from Section \ref{sec:MOR} that we can construct a set of POD bases whose linear combination spans the finite dimensional concentration field $c_d$ of the AD model \eqref{eq:VBVP} as
$$ c_d(\bbx) = \sum_{i=1}^N c_i \, \psi_i(\bbx) = \bbpsi(\bbx) \, \bbc ,$$
where $\bbpsi = [\psi_1 \ \dots \ \psi_N]$ and $\bbc \in \reals^N$.
Given a set of values for the source parameters $\bbp$, we can calculate the coefficients for this linear representation as $\bbc = \bbA^{-1} \bbb(\bbp)$, where matrix $\bbA$ and vector $\bbb$ are defined in equation \eqref{eq:VBVPLS} in Appendix \ref{sec:Grad}. Thus
$ c_d(\bbx; \bbp) = \bbpsi (\bbx) \, \bbA^{-1} \bbb(\bbp) .$
Therefore, we can calculate the desired derivative in the definition of the FIM \eqref{eq:defFIM} as
$$ \frac{\partial c_d(\bbx, \bbp)}{\partial \bbp} = \bbpsi (\bbx) \, \bbA^{-1} \nabla_{\bbp} \bbb ,$$
where $\nabla_{\bbp} \bbb = - \bbM_{\bbp}$ and $\bbM_{\bbp}$ is the derivative of the finite dimensional AD model \eqref{eq:Model} with respect to $\bbp$ and is given via equation \eqref{eq:Mp} in Appendix \ref{sec:secOrderInfo}. Let
\begin{equation} \label{eq:Sp}
\bbS(\bbp) \triangleq \bbA^{-1} \bbM_{\bbp} ,
\end{equation}
be a function of $\bbp$ with $\bbS \in \reals^{N \times p}$, and without loss of generality assume $\bbarsigma_{\max} = 1$. Note that as long as the variance of the noise is constant, its value is irrelevant for planning. Then, from equation \eqref{eq:defFIM}, we get
$ \bbF(\tbx_m; \bbp) = \sum_{i=1}^m \bbS(\bbp)^T \bbpsi(\bbx^i)^T \bbpsi(\bbx^i) \bbS(\bbp)  $
or in matrix form
\begin{equation} \label{eq:FIM}
\bbF(\tbx_m; \bbp) = \bbS(\bbp)^T \bbX(\tbx_m)^T \bbX(\tbx_m) \, \bbS(\bbp),
\end{equation}
where
\begin{equation} \label{eq:X}
\bbX(\tbx_m) = 
\left[
\begin{array}{ccc}
  \psi_1(\bbx^1) &   \dots	& \psi_N(\bbx^1)   \\
  \vdots 		 &  \ddots &   \vdots   \\
  \psi_1(\bbx^m) &  \dots 		&   \psi_N(\bbx^m)
\end{array}
\right] ,
\end{equation}
is the $m \times N$ design matrix.

Given the sequence of waypoints $\tbx_m = (\bbx^1, \dots, \bbx^m)$ at step $m>\bbarm$ and the corresponding vector of noisy measurements $\bby_m$, we solve the SI problem \eqref{eq:optFD} to obtain the estimation $\bbp_m$ of the unknown source parameters at current step.
Then, the Path Planning problem consists of finding the next best waypoint $\bbx^{m+1}$ from where if a new measurement is taken, it will maximize the minimum eigenvalue of the FIM. In mathematical terms
\begin{equation} \label{eq:pathOptL}
\bbx^{m+1} = \argmax_{\bbx \in \Omega} \lambda_{\min} [ \bbF_m + \bbS(\bbp_m)^T \bbpsi(\bbx)^T \bbpsi(\bbx) \, \bbS(\bbp_m) ] ,
\end{equation}
where $\bbF_m = \bbF(\tbx_m, \bbp_m) \in \mbS_+^p$ is a constant FIM, defined by equation \eqref{eq:FIM}, that contains the information from the current $m$ measurements. The second term in the right-hand-side of \eqref{eq:pathOptL} is a rank-one update capturing the information added by measuring at a new location $\bbx$.
Given the solution of \eqref{eq:pathOptL}, we use a motion planner to navigate the robot from its current position to the next measurement location $\bbx^{m+1}$ while avoiding obstacles; see Section \ref{sec:experiment} for more details.
Note that since $\Omega \subset \reals^d$, the dimension of \eqref{eq:pathOptL} is very small which makes it particularly attractive for online implementation on a mobile robot.
The proposed planning algorithm is presented in Algorithm \ref{alg:pathPlan}.
\begin{algorithm}[t]
\caption{Optimal Waypoint Selection Algorithm}
\label{alg:pathPlan}
\begin{algorithmic}[1]

\REQUIRE The POD bases $\bbpsi = [\psi_1, \dots, \psi_N]$ of Algorithm \ref{alg:POD};

\REQUIRE The number of initial measurements $\bbarm$ and the maximum number of measurements $m_{\max}$;

\STATE  Collect the initial measurements and set $\tbx_{\bbarm} = (\bbx^1, \dots, \bbx^{\bbarm})$ and $\bby_{\bbarm} = (c^m(\bbx^1), \dots, c^m(\bbx^{\bbarm}))$;

\FOR{ $m = \bbarm \ \textbf{to} \ m_{\max}$ }

\STATE Solve the SI problem \eqref{eq:optFD} with $\bby_m$ to get $\bbp_m$;

\STATE Compute $\bbS_m = \bbS(\bbp_m)$ and the design matrix $\bbX_m = \bbX(\tbx_m)$ according to equations \eqref{eq:Sp} and \eqref{eq:X};

\STATE Compute the constant matrix $\bbF_m  = \bbS_m^T \bbX_m^T \bbX_m \bbS_m$;

\STATE Given $\bbS_m$ and $\bbF_m$, solve the planning problem \eqref{eq:pathOptL} for $\bbx^{m+1}$ utilizing the SSDP approach of Algorithm \ref{alg:SSDP};			\label{line:SSDP}

\STATE Update the waypoints $\tbx_{m+1} = (\tbx_m, \bbx^{m+1})$;

\STATE Update $\bby_{m+1} = (\bby_m, c^m(\bbx^{m+1}) )$ from equation \eqref{eq:measureFun};

\STATE $m \leftarrow m+1$;

\ENDFOR

\end{algorithmic}
\end{algorithm}
Line \ref{line:SSDP} corresponds to solving the Next Best Measurement Problem \eqref{eq:pathOptL}. To do so, we reformulate \eqref{eq:pathOptL} into a Semi-Definite Programming (SDP) problem and solve it using the Sequential SDP (SSDP) method; see Appendix \ref{sec:SSDP} for details.

\section{Active Source Identification\\ in Complex Domains} \label{sec:autoSI}
The developments of Sections \ref{sec:SI} and \ref{sec:path} relied on the assumption that the domain of interest $\Omega$ is represented by a set of convex box constraints. Although extension to handle a set of affine constraints defining $\Omega$ or even any other set of convex constraints is straightforward, this is not the case if $\Omega$ is non-convex.
To solve the SI problem discussed in Section \ref{sec:SI} in a non-convex domain $\Omega$, we first decompose this domain into convex subdomains. Then, using the Sensitivity Analysis (SA) initialization method discussed in Appendix \ref{sec:init}, we select the largest subdomains that contain the candidate source locations and solve the SI problem \eqref{eq:optFD} in those subdomains.
To solve the planning problem discussed in Section \ref{sec:path} in non-convex domains, we follow a similar approach. Particularly to determine every new waypoint $\bbx^{m+1}$ of the robot, we define a subdomain of $\Omega$ around the initialization point $\bbx_0$, given by equation \eqref{eq:initialize} in Appendix \ref{sec:pathInit}, and solve the nonlinear SDP \eqref{eq:NLSDP} in this region.
Note that this initialization scheme ensures existence of a local optimum in the selected subdomain and  preserves the global convergence property of the SSDP Algorithm \ref{alg:SSDP} for non-convex domains.

\begin{figure}[t!]
  \centering
    \includegraphics[width=0.45\textwidth]{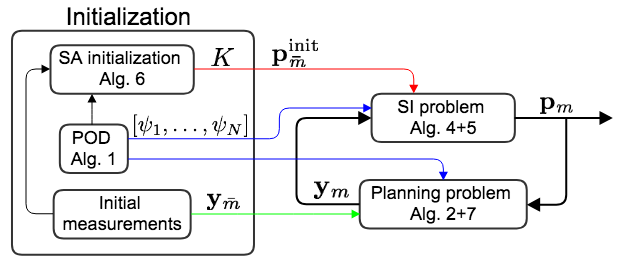}
        \caption{Schematic representation of the ASI Algorithm \ref{alg:autoSI}. After collection of $\bbarm$ initial measurements, the robot solves the SI problem \eqref{eq:optFD} and path planning problem \eqref{eq:pathOptL} sequentially and in a feedback loop for $m > \bbarm$.} \label{fig:autoSI}
\end{figure}

Integrating the solution of the SI problem with the planning problem, discussed in Sections \ref{sec:SI} and \ref{sec:path}, respectively, in a feedback loop and incorporating the proposed domain decomposition method to handle optimization in non-convex domains, we obtain the proposed Active Source Identification (ASI) method.
Specifically, given a set of initial measurements, the robot solves the SI problem as discussed in Section \ref{sec:SI} over the subdomains that contain possible source locations as indicated by the SA method in Algorithm \ref{alg:init}. Then, given the solution of the SI problem, the robot plans its next measurement according to the procedure developed in Section \ref{sec:path}, and the process repeats. The proposed ASI algorithm terminates when
\begin{equation} \label{eq:sutoStop}
\norm{\bbp_m - \bbp_{m-1}}_2 \leq  \epsilon
\end{equation}
for some $0 < \epsilon \ll 1$, where $\bbp_m$ is the solution of the SI problem \eqref{eq:optFD} at step $m$.
The proposed integrated method is summarized in Algorithm \ref{alg:autoSI} and illustrated in Figure \ref{fig:autoSI}.
\begin{algorithm}[t!]
\caption{Active Source Identification}
\label{alg:autoSI}
\begin{algorithmic}[1]

\REQUIRE The stopping tolerance $\epsilon$ of equation \eqref{eq:sutoStop};
	
\STATE Take initial measurements to get $\bby_{\bbarm}$;

\STATE Given $E_{\bbarm}$ corresponding to $\bby_{\bbarm}$, utilize the SA Algorithm \ref{alg:init} from Appendix \ref{sec:init} to initialize the SI problem. 

\STATE Decompose the domain $\Omega$ into subdomains containing $K$ cluster centers; 	 \label{line:Er}

\STATE Set $m \leftarrow \bbarm$;

\WHILE{the algorithm has not converged}

\STATE Solve the SI problem \eqref{eq:optFD} for source parameters $\bbp_m$, initialized by $\bbp_{m-1}$, using the results of Section \ref{sec:SI} with $E_m$ corresponding to $\bby_m$;	 \label{line:init}

\STATE Check the convergence criterion \eqref{eq:sutoStop};

\STATE Take a new step using Algorithm \ref{alg:pathPlan} and given $\bbp_m$;

\STATE Update the measurement vector $\bby_{m+1}$;

\STATE $m \leftarrow m+1$;

\ENDWHILE

\end{algorithmic}
\end{algorithm}
In lines \ref{line:Er} and \ref{line:init} of Algorithm \ref{alg:autoSI}, $E_m$ denotes the set of infinitesimal areas enclosing measurement locations; cf. Section \ref{sec:probDef}.
Note that via successive solutions of the SI problem in line \ref{line:init}, the solver corrects its previous estimates of the source parameters $\bbp_m$ taking into account the most recent concentration measurement $y_m$.
The initialization of the SI problem with the previous solution results in faster convergence.

\section{Results} \label{sec:sim}
In this section we provide numerical simulations and real-world experiments to illustrate the ASI Algorithm \ref{alg:autoSI}. 
We solve the constrained nonlinear optimization problem (\ref{eq:optFD}) utilizing the \texttt{fmincon($\cdot$)} function in the \textsc{MATLAB} optimization toolbox that employs an interior-point algorithm which accepts the Hessian-vector multiplication information; cf. Appendix \ref{sec:secOrderInfo}.
We also use the \texttt{clusterdata($\cdot$)} function in \textsc{MATLAB} to perform the clustering required in Algorithm \ref{alg:init} of Appendix \ref{sec:init}.
Moreover, we solve the SDP \eqref{eq:tangent} with a primal-dual method using \textsc{cvx}, a package for specifying and solving convex programs \cite{CVX}. For more details about interior-point and primal-dual algorithms, see, e.g., \cite{NO2006NW}.

In order to quantify the performance of our method, we report four different error metrics, namely the uncovered source ratio $e_{\text{un}}$, the false detection ratio $e_{\text{fd}}$, the normalized intensity error $e_{\text{int}}$, and the normalized localization error $e_{\text{loc}}$.
In mathematical terms
$ e_{\text{un}} = { \norm{\bbars_d - s_d}_{\chi_F} }/{ \norm{\bbars_d}_{L^2} } \ \ \text{and} \ \ 
e_{\text{fd}} = { \norm{s_d}_{\chi_{\Omega \setminus F}} }/{ \norm{\bbars_d}_{L^2} } , $
where $F$ is the support set of the true source $\bbars_d$, $\chi_F$ denotes the indicator function of $F$ defined in equation \eqref{eq:indicFun}, and $\norm{\cdot}_{\chi_{F}}$ is introduced in \eqref{eq:Xnorm}.
The error term $e_{\text{un}}$ measures the fraction of the true source $\bbars_d$ that is left out by the estimated source $s_d$ and the error term $e_{\text{fd}}$ considers the parts of the estimated source $s_d$ that do not overlap with the true source $\bbars_d$. Note that any value $e_{\text{un}} < 1$ indicates an overlap between the true and estimated sources.
Finally, the errors $e_{\text{int}}$ and $e_{\text{loc}}$ are defined for a single source as $e_{\text{int}} = \abs{\bbarbeta - \beta} / \beta_{\max}$ and $e_{\text{loc}} = \norm{\barbz - \bbz}_2 / l$, where $\beta_{\max}$ is an upper bound on the source intensity, $\bbz \in \Omega$ is the center of the rectangular source support, and $l$ is the characteristic length of the domain $\Omega$.
We also define the signal to noise ratio for simulations in dBs as
$ \text{SNR} = 20 \log \left( {\norm{c^m (\bbx)}_{\chi_E}} / {\norm{\epsilon(\bbx)}_{\chi_E}} \right) , $
where $\epsilon(\bbx)$ denotes the noise field.

We study the performance of the ASI Algorithm \ref{alg:autoSI} as a function of the dimensionless Peclet number, which is a measure of the relative dominancy of advection over diffusion and is defined as $Pe = {u \, l}/{\bbarkappa}$, where $l$ is the characteristic length, $u$ is the magnitude of the inlet velocity, and $\bbarkappa$ is the average diffusivity of the medium.

To generate the snapshots for POD Algorithm \ref{alg:POD}, and to solve problem \eqref{eq:optFD} numerically, we need to obtain the solution of AD model \eqref{eq:BVP} for a given source function. To this end, we discretize the domain $\Omega$ and use the FE method with standard Galerkin scheme \cite{FCFEM2007L}.
Let $n$ denote the size of the required FE mesh which we generate using \textsc{cubit} \cite{cubit}. 
We construct the discrete FE models using an in-house FE code based on the DiffPack \textsc{C++} library \cite{CPDENMDP2003L}.
Moreover, to approximate the first- and second-order derivatives over FE meshes, we use finite difference for structured meshes and polynomial interpolation for unstructured meshes.

We select the thresholding parameter of the SA Algorithm \ref{alg:init} as $\alpha = 0.7$; see Appendix \ref{sec:init}.
To build the POD basis functions via Algorithm \ref{alg:POD}, we need to generate snapshots of the AD-PDE \eqref{eq:BVP}. Since the relationship between the magnitudes of the source and concentration functions is linear, cf. equation \eqref{eq:ADPDE}, we cover the domain $\Omega$ with tower functions \eqref{eq:towerF} with unit intensity for this purpose.
Finally, we set the regularization parameter in \eqref{eq:obj} to $\tau = 10^{-8}$ and the stopping tolerance in the ASI Algorithm \ref{alg:autoSI} to $\epsilon = 10^{-3}$.

\subsection{Numerical Simulations} \label{sec:ppComp}
\begin{figure}[t!]
  \centering
    \includegraphics[width=0.5\textwidth]{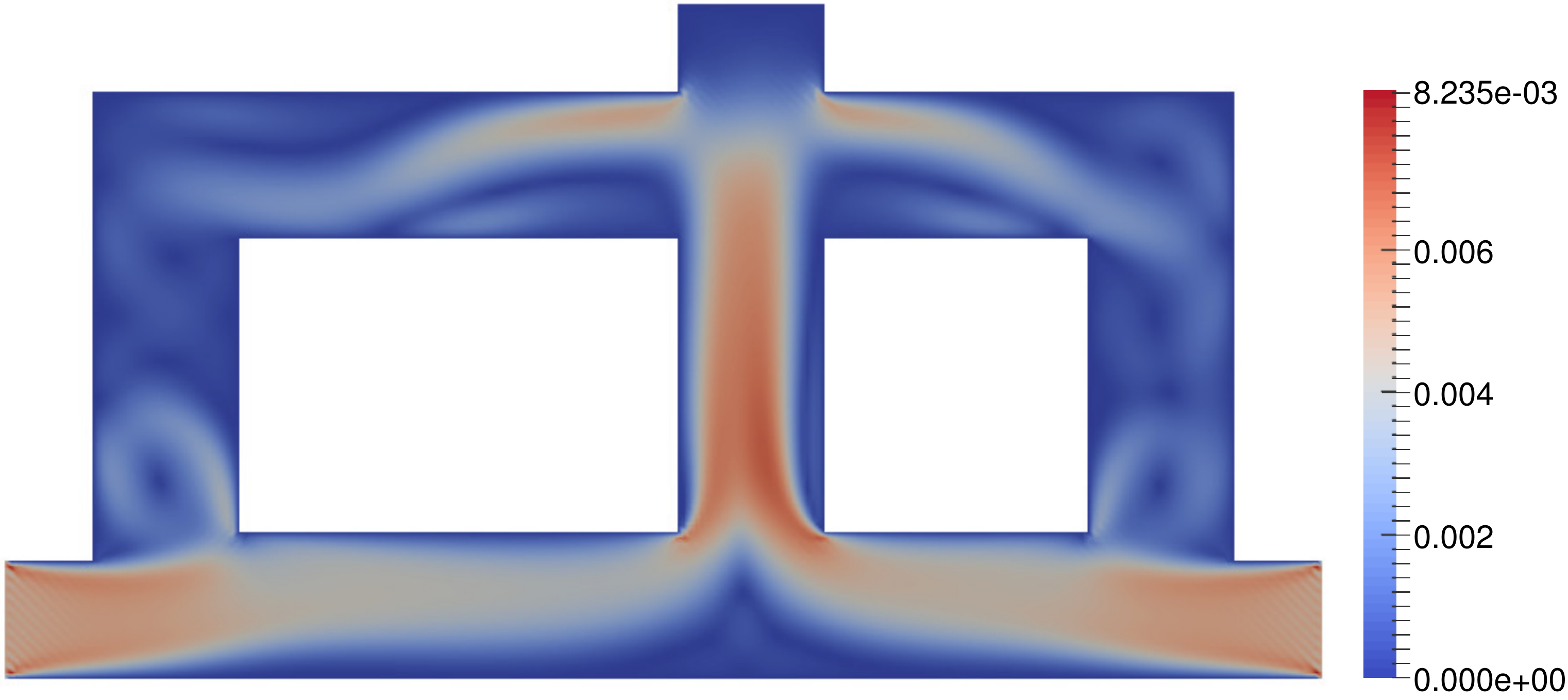}
        \caption{Steady-state flow pattern for the non-convex domain. } \label{fig:Ex_BI-flow}
\end{figure}

In this section, we study the performance of the ASI Algorithm \ref{alg:autoSI} in a large non-convex domain $\Omega$. 
We assume that the air flows into the domain through the sides, i.e., constant velocity inlet boundary conditions. Then, an in-house fluid dynamics code is utilized to simulate the steady-state velocity in the domain as depicted in Figure \ref{fig:Ex_BI-flow}, where a FE mesh with $n = 15034$ points is used.
We consider Peclet numbers $Pe = 2.5, Pe = 25$, and $Pe = 250$. Moreover to reduce the order of the model, we utilize $R = 597$ snapshots with $\eta = 0.97$ corresponding to $N = 154, N = 183$, and $N = 205$ basis functions for each Peclet number, respectively.

In our first simulation study, we compare the planning method presented in Section \ref{sec:path} to placement over a lattice and an ergodic placement method that uses the determinant of the FIM as the information metric.
Both of these planning approaches employ the algorithm developed in Section \ref{sec:SI} to solve the SI problem \eqref{eq:opt}
%
and, therefore, the purpose of this comparison is to showcase the relative performance of the planning method proposed here.
Specifically, given an information distribution, the ergodic planner aims at designing paths where the time spent by the robot at any region is proportional to the information at that region. This approach is proposed in \cite{MEDEDMAS2011MM} and used for active sensing in \cite{EEDI2016MSMM}. Specifically, we use the normalized determinant of the FIM \eqref{eq:FIM} as the information distribution in the ergodic planner, where we assign zero information value to the points on obstacles. We implement the controller proposed in \cite{MEDEDMAS2011MM} and, similar to the ASI Algorithm \ref{alg:autoSI}, we update the information distribution at every step as newer estimates of the source parameters become available.

To highlight the advantages of the proposed ASI framework against the heuristic methods discussed in Section \ref{sec:introSI}, we also compare our algorithm to a heuristic approach that drives the robot along the normalized concentration gradient ascent and upwind directions, as proposed in \cite{OMRN2002MND}. The robot uses the initial $\bbarm$ measurements to detect the plume and initializes its path from the highest measured concentration point. The velocity field is known exactly to the robot and the concentration gradient at each point is approximated by taking two additional measurements in orthogonal directions. Since the heuristic approach only provides a location estimate, it is compared to the other methods in terms of $e_{\text{loc}}$.

\begin{figure}[t!]
  \centering
    \includegraphics[width=0.5\textwidth]{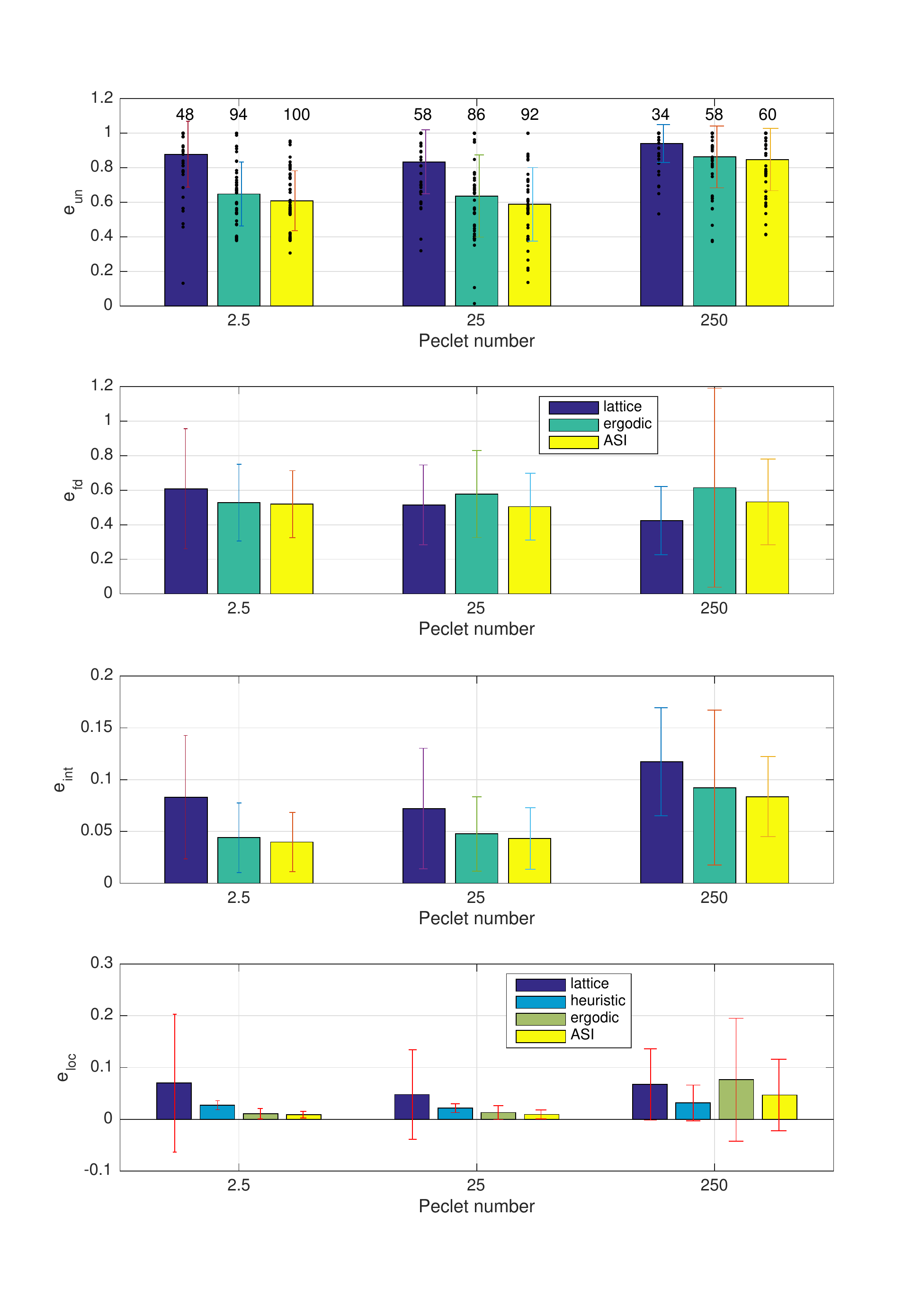}
        \caption{Comparison of the ASI Algorithm \ref{alg:autoSI} with lattice and ergodic placements, and heuristic approach for three Peclet numbers. The figure depicts the average uncovered, false detection, intensity, and localization errors, respectively. The standard deviation bars and scatter plots are overlaid on the error bars. The success percents are also given on top of the first subfigure.} \label{fig:Ex_BI-LATvsEXP}
\end{figure}

In the following simulations, we use $\bbarm = 28$ initial measurements for the ASI, ergodic, and heuristic methods and set the maximum number of steps to $m_{\max} = 42$. The lattice placement uses $m_{\max} = 42$ measurements obtained by sensors located on an equidistant grid. For the ergodic and heuristic methods, we use a first-order model for the dynamics of the robot.
The results are plotted in Figure \ref{fig:Ex_BI-LATvsEXP} where we average over $50$ randomly generated sources. 
It can be seen that the proposed planning Algorithm \ref{alg:pathPlan} outperforms the lattice method in all cases and performs more consistently in terms of standard deviation. Particularly, the success rate of the ASI algorithm, i.e., the number of instances that the algorithm finds an overlapping source estimate, is considerably higher for all three Peclet numbers.
Note that the average false detection error of the lattice placement for $Pe = 250$ is smaller than the ASI algorithm but since this method often fails to find an overlapping source estimate for this Peclet number, $e_{\text{fd}}$ only indicates that the falsely detected sources have smaller volumes than the true sources on average.

The performance of the ergodic approach is close to the proposed planning method since it uses a similar optimality index to collect the measurements. Note that as the domain becomes larger, the computation of the information distribution required by this approach becomes expensive rendering this planning method intractable.
Furthermore, since the performance of the ergodic method depends on the combination of exploration and exploitation \cite{EEDI2016MSMM}, we allow the robot to travel through obstacles and take a measurement every three steps so that it can reach more informative regions of the domain more often.
To the contrary, the behavior of the proposed ASI method indicates that given an initial set of measurements, necessary to detect the unknown sources, the most informative measurements are obtained close to the location of sources as opposed to points farther away. Therefore, the better performance of the ASI algorithm, i.e., its smaller and more consistent false detection error values, can be attributed to this fact.

Considering the last subfigure in Figure \ref{fig:Ex_BI-LATvsEXP}, we observe that the ASI algorithm provides more accurate localization for $Pe=2.5$ and $Pe=25$ but the heuristic method performs better for $Pe=250$.
The reason for this is that for very high Peclet numbers, for which advection is the primary means of transport, the reduced order AD model \eqref{eq:Model} becomes inaccurate resulting in poor localization for model-based methods; see Section \ref{sec:discussion}.
Nevertheless, the heuristic approach does not provide any information about the size or intensity of the source and in the case of multiple sources, it localizes at most one source or fails altogether.

\begin{figure}[t] 
        \centering
        \begin{subfigure}[b]{0.5\textwidth}
                \includegraphics[width=\textwidth]{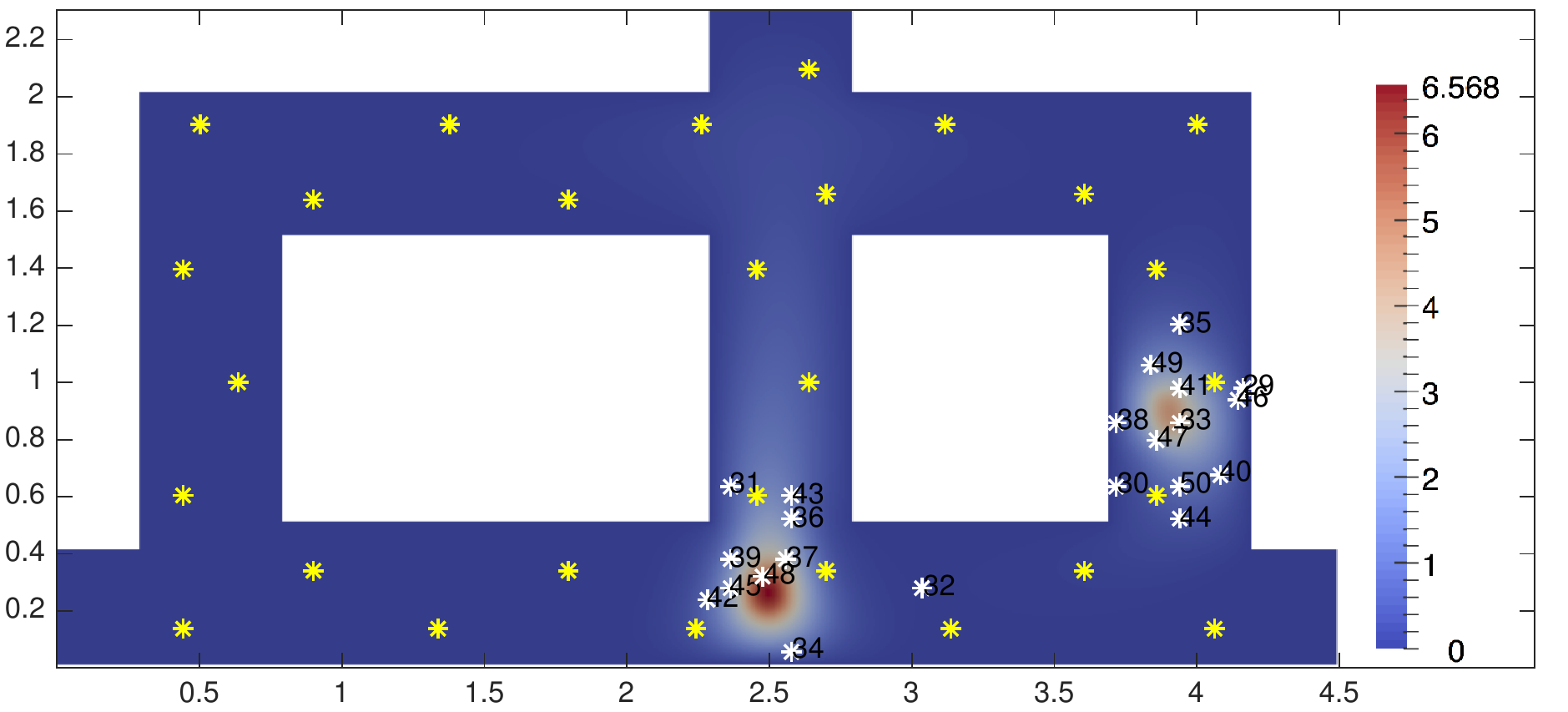}
    \caption{waypoints of the mobile robot} \label{fig:ExBII_path}
    \end{subfigure}
    \quad 
        \begin{subfigure}[b]{0.5\textwidth}
                \includegraphics[width=\textwidth]{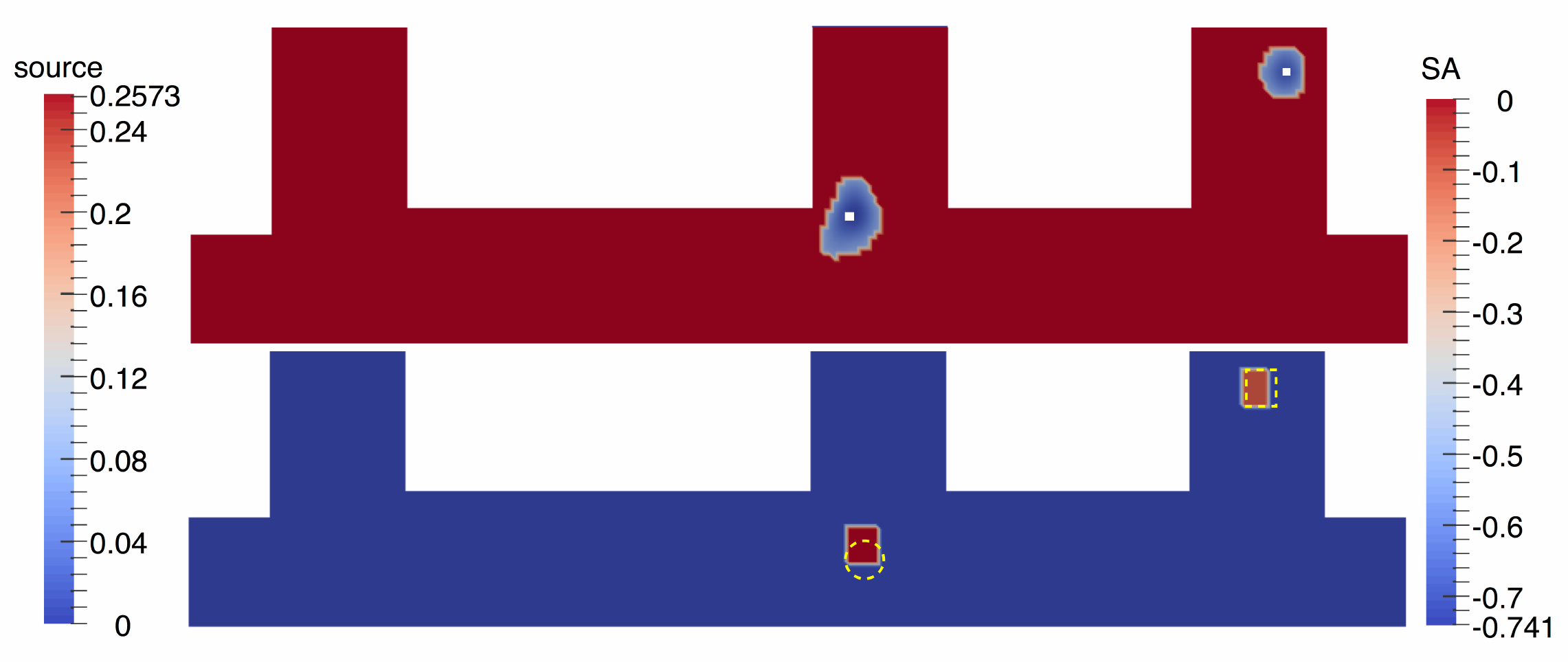}
    \caption{estimated source} \label{fig:ExBII_source}
    \end{subfigure}
        \caption{Waypoints of the mobile robot overlaid on the concentration field and the contours of the SA initialization and estimated source. The yellow stars in Figure \ref{fig:ExBII_path} indicate the initial measurements while the white stars show the sequence of waypoints. In Figure \ref{fig:ExBII_source}, the SA initialization (top) and the final solution (bottom) are shown, where the white squares depict the support of the initial estimate and the yellow lines delineate the support of the true source.} \label{fig:Ex_BII-twin}
\end{figure}

In our second simulation study, we use the same settings as before and consider an AD transport with $Pe = 25$ and two sources, specifically, a circular source centered at $(2.5, 0.25)$ with radius of $0.08$ and intensity of $0.25$ and a rectangular source with parameters $\barbp = (0.2, 3.85, 3.95, 0.8, 0.95)$ creating the concentration filed given in Figure \ref{fig:ExBII_path}.
Since the two sources are not located in one convex domain, decomposing $\Omega$ into convex subdomains and following the procedure described in Section \ref{sec:autoSI} is necessary to recover both sources. We note that the ASI algorithm has no \textit{a priori} knowledge of the number of sources.
It solves the problem in $1951$sec in $22$ steps, which amounts to solving $22$ instances of the SI problem \eqref{eq:optFD}. Time required to solve the planning problem \eqref{eq:pathOptL} is negligible. The final error values are $e_{\text{un}} = 0.67$ and $e_{\text{fd}} = 0.61$ with SNR $= 19.02$ dB.
The waypoints of the robot are given in Figure \ref{fig:ExBII_path} by white stars. Note the accumulation of the measurements around the high concentration regions of the domain, i.e., the hot spots \cite{MIPPASEP2014CLD}.
Note also that to cover both sources simultaneously, the robot needs to move back and forth between them. We can minimize the travelled distance, by adding a penalty term in the planning problem \eqref{eq:pathOptL} to encourage more measurements before moving to the next source but this would be suboptimal from an information perspective. A more viable option is to use multiple robots, which is part of our future work.
The result of the SA initialization Algorithm \ref{alg:init} and the final solution with the true source overlaid on it are plotted in Figure \ref{fig:ExBII_source}.
%
Note that the performance of the ASI Algorithm \ref{alg:autoSI} only depends on the dimensionless Peclet number. The units of the other quantities are arbitrary as long as they are selected consistently. Particularly, given a unit for concentration $c$, the unit for source term $s$ is defined as concentration per unit time; cf. AD-PDE \eqref{eq:BVP}. See the next section for a specific example.

\subsection{Experimental Results} \label{sec:experiment}
\begin{figure}[t!]
  \centering
    \includegraphics[width=0.4\textwidth]{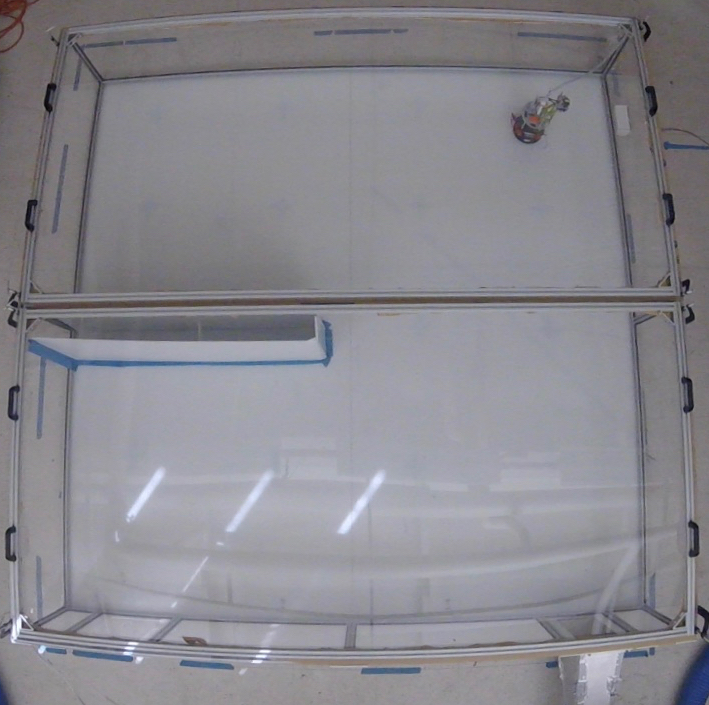}
        \caption{Non-convex domain of the experiment. The velocity inlet, velocity outlet, and desired source are located in the lower-right, lower-left, and upper-right corners, respectively.} \label{fig:exp-dom}
\end{figure}

%
In this section, we demonstrate the performance of the proposed ASI Algorithm \ref{alg:autoSI} experimentally for the identification of an ethanol source in air. Particularly, we consider a non-convex domain with dimensions $2.2 \times 2.2 \times 0.4$m$^3$, cf. Figure \ref{fig:exp-dom}.
We connect a fan to the domain through a duct that blows air into domain with an axial speed of $0.9$m/s and a tangential speed of $0.2$m/s, creating a turbulent flow. We utilize \textsc{ANSYS-Fluent} to obtain the desired flow properties namely, the velocity and diffusivity fields. Since turbulence is a $3$D phenomenon, we build a $3$D mesh of the domain with $1.94 \times 10^6$ elements.
We note that determining the velocity and diffusivity fields for turbulent flow is non-trivial.
Turbulent flow is recognized with high Reynolds numbers and is characterized by severe fluctuations in the flow properties. These fluctuations enhance the mixing in the flow and facilitate the transport of the concentration. This enhanced mixing is often encapsulated in an effective turbulent diffusivity which is proportional to effective turbulent viscosity $\mu$ and the proportionality constant is the Schmidt number Sc. Then, the total diffusivity is the sum of laminar $\kappa_0$ and turbulent diffusivities \cite{TD2002RW}. Mathematically,
\begin{equation} \label{eq:diff}
\kappa = \kappa_0 + \frac{\mu}{\rho \, \text{Sc}} ,
\end{equation}
where $\rho \in \reals_+$ denotes the density of the transport medium.

\begin{figure}[t] 
        \centering
        \begin{subfigure}[b]{0.4\textwidth}
                \includegraphics[width=\textwidth]{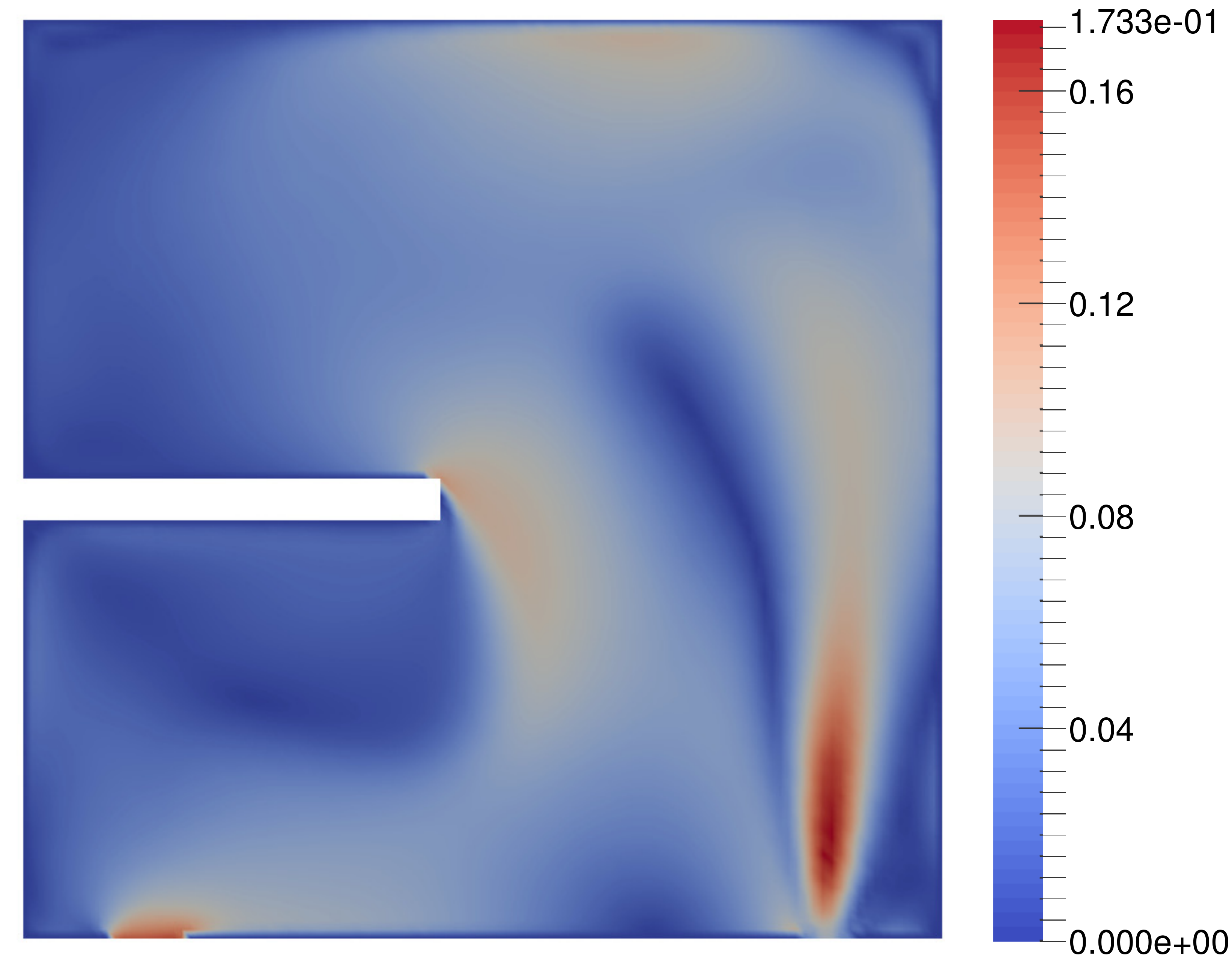}
    \caption{velocity field (m/s)} \label{fig:exp-vel}
    \end{subfigure}
    \quad 
        \begin{subfigure}[b]{0.4\textwidth}
                \includegraphics[width=\textwidth]{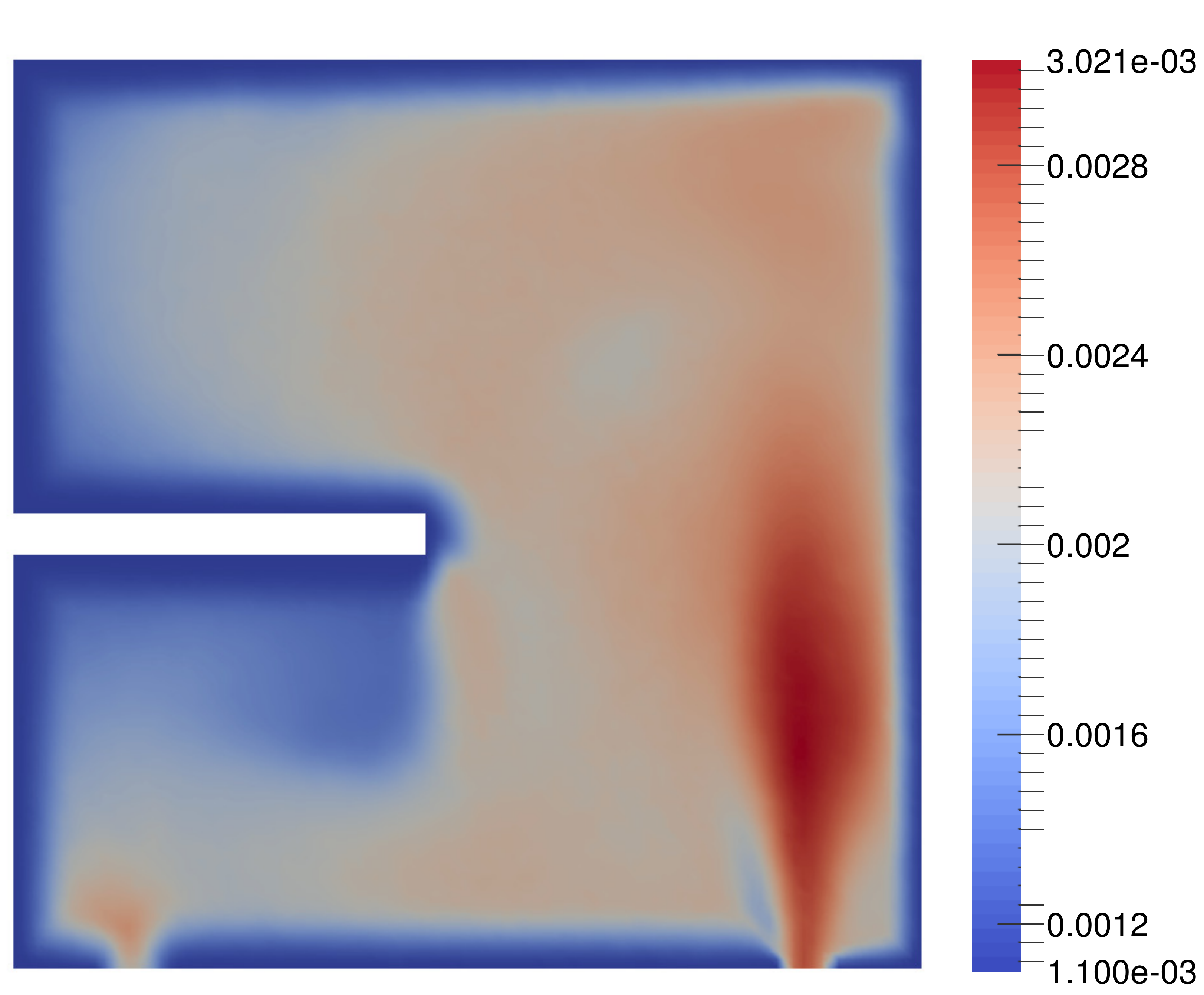}
    \caption{diffusivity field (m$^2$/s)} \label{fig:exp-diff}
    \end{subfigure}
        \caption{Turbulent flow properties required for the solution AD-PDE. The fields are interpolated to the plane of concentration sensor located at a height of $0.27$m. Top figure shows the predicted velocity field generated by blowing air through the inlet using a fan. The bottom plot shows the corresponding total diffusivity field which is the sum of laminar and turbulent diffusivities, given by \eqref{eq:diff}, and lower bounded by $10^{-3}$m$^2$/s to stabilize the AD FE model.} \label{fig:exp-turbSol}
\end{figure}

We construct a $2$D discretized AD model using the FEM with $n=12121$ grid points located on the plane of the robot concentration sensor at height of $0.27$m. Using the $2$D model instead of a $3$D model is an approximation since we ignore the transport in $x_3$-direction but considerably decreases the computational cost.
The turbulent flow properties at the nodes of the AD FE mesh are interpolated from the \textsc{ANSYS-Fluent} $3$D model and are given in Figure \ref{fig:exp-turbSol}. The corresponding Peclet number is Pe $= 213$.
Regarding the diffusivity field, the following points are relevant. For molecular diffusion of ethanol in air, $\kappa_0 \approx 1.1 \times 10^{-5}$m$^2$/s. Then, from Figure \ref{fig:exp-diff} it can be seen that the turbulent diffusivity is considerably larger than the laminar diffusivity. This contributes to the numerical stability of the AD model by decreasing the Peclet number.
To further increase the stability, we add an artificial diffusion lower bounding the total diffusivity \eqref{eq:diff} by $10^{-3}$m$^2$/s; see the discussion of Section \ref{sec:discussion} for more details.
Given the discretized model, we use $N = 900$ basis functions to construct the reduced model \eqref{eq:Model}.

\begin{figure}[t!]
  \centering
    \includegraphics[width=0.2\textwidth]{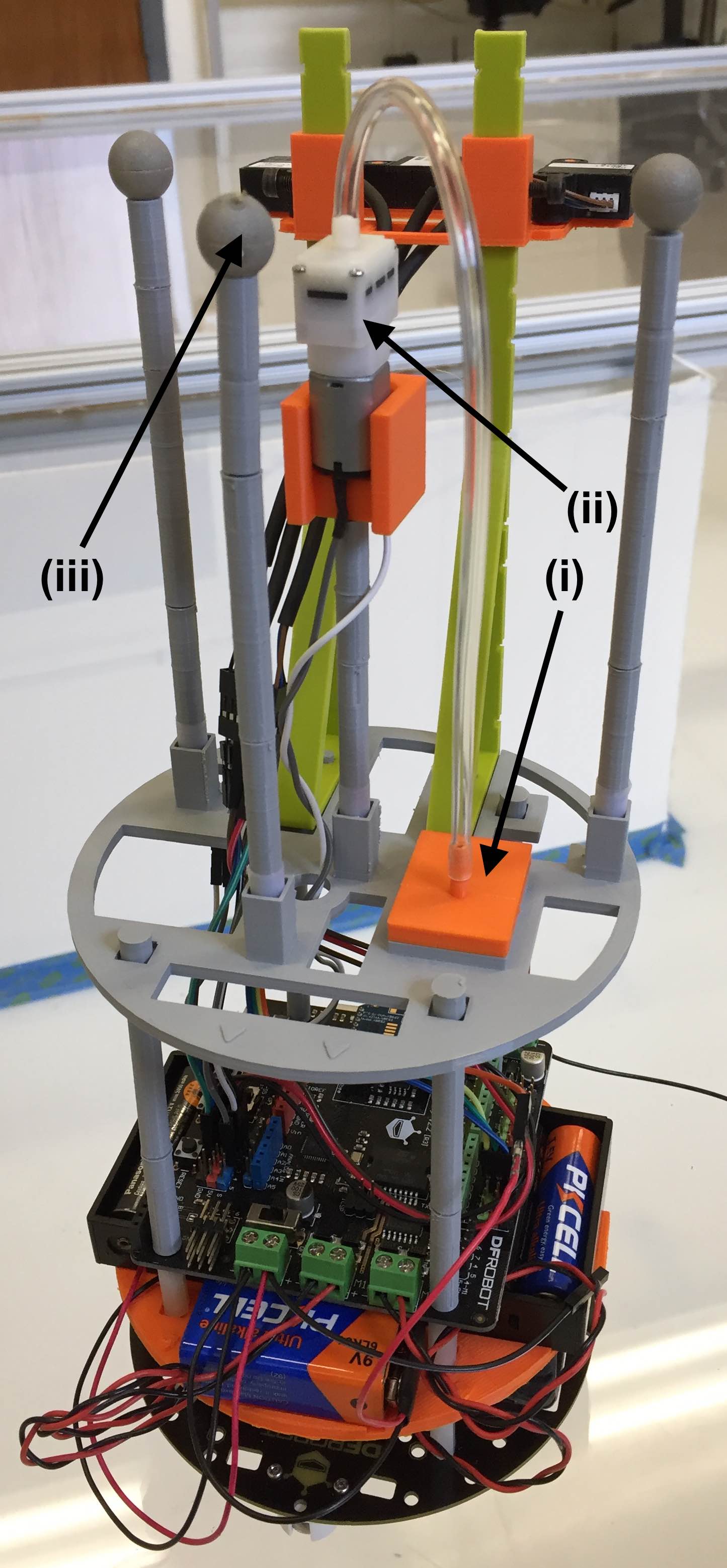}
        \caption{Mobile robot used to collect the measurements: (i) the concentration sensor placed in a confined box to separate it from flow conditions, (ii) air pump, (iii) \textsc{OptiTrack} markers used for localization. The robot is remotely controlled by a computer via radio communication.} \label{fig:exp-robot}
\end{figure}

The ethanol source is located at $x_1 = 1.8$m, $x_2 = 1.8$m, and $x_3 = 0.3$m, across from the velocity inlet and releases ethanol at a steady rate. To collect the measurements, we use a custom built differential drive mobile robot equipped with a \textsc{MiCS-5524} concentration sensor. In order to eliminate the effect of velocity field on the sensor readings, we place the sensor in a confined box and utilize an air pump to deliver air to the sensor with an approximately constant flow rate, cf. Figure \ref{fig:exp-robot}.
As mentioned in Section \ref{sec:introSI}, non-smooth concentration patches appear when the flow is turbulent. To ensure the detection of these patches, we need to allow enough instantaneous readings at each location. Furthermore, to minimize the effect of intervals of low concentration between the detections, we calculate the final concentration value as the average of the readings at the highest quartile. In the experiment, we record $1000$ instantaneous readings at each measurement location.
We perform the computations off-board and communicate the commands to the robot via radio communications. This allows us to use a very small robot minimizing the interference with the flow field. The localization needed for motion control of the robot, is provided by an \textsc{OptiTrack} motion capture system and a simple controller is implemented for line tracking. We utilize the \textsc{VisiLibity} toolbox to generate the geodesic path between each pair of waypoints given by the planning Algorithm \ref{alg:pathPlan}, taking into account the obstacle; see \cite{VisiLibity2008O}.

\begin{figure}[t] 
        \centering
        \begin{subfigure}[b]{0.48\textwidth}
                \includegraphics[width=\textwidth]{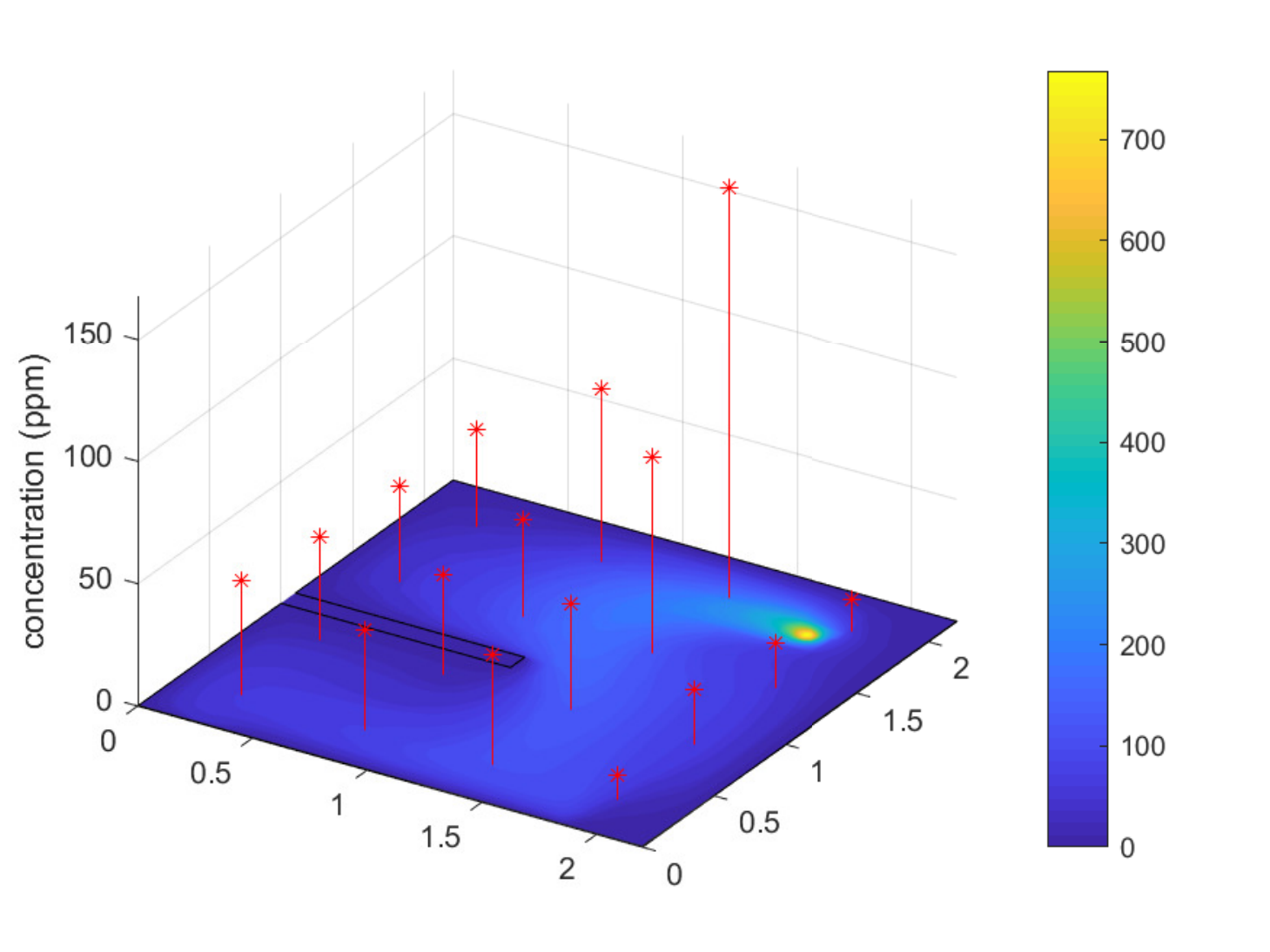}
    \caption{initial measurements} \label{fig:exp-meas}
    \end{subfigure}
    \quad 
        \begin{subfigure}[b]{0.4\textwidth}
                \includegraphics[width=\textwidth]{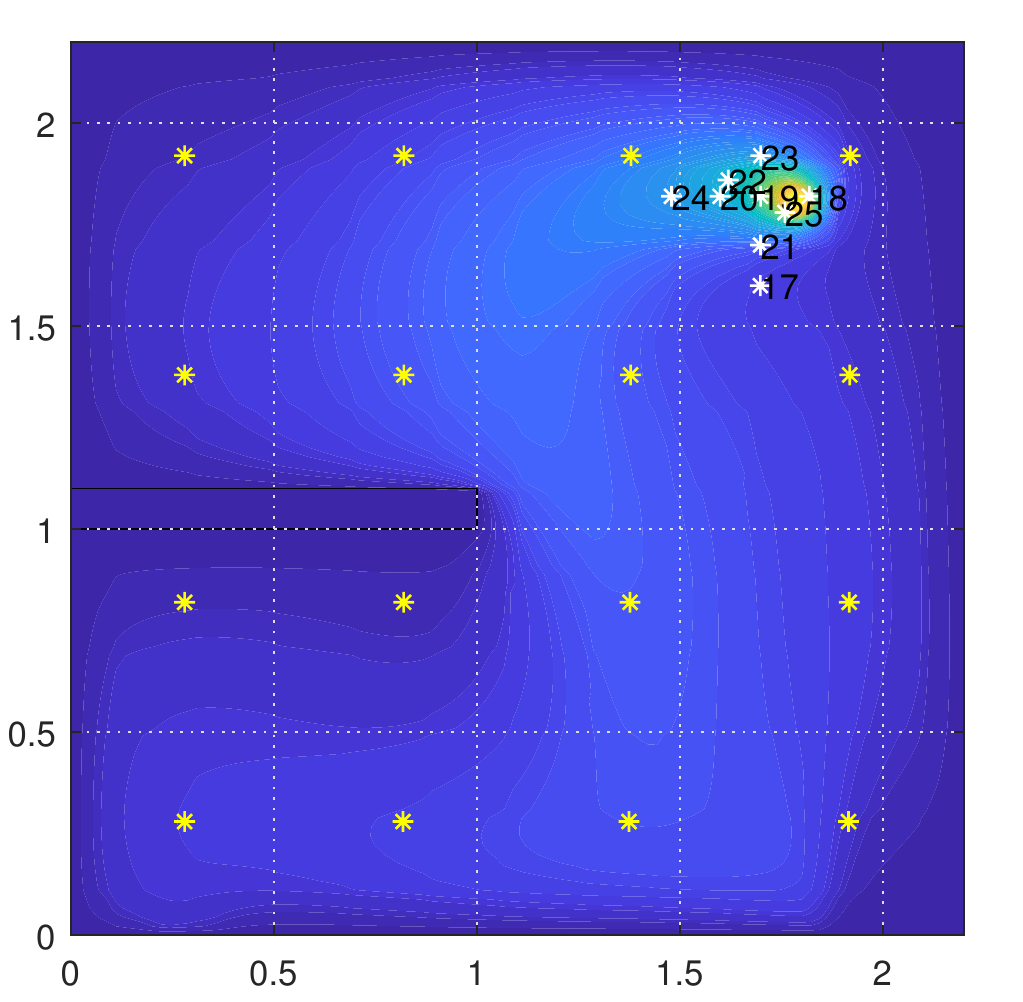}
    \caption{waypoints} \label{fig:exp-place}
    \end{subfigure}
        \caption{Top figure shows the initial $\bbarm$ measurements overlaid on the concentration field predicted by AD-PDE \eqref{eq:BVP} for a hypothetical source located at the true location. Bottom figure shows the waypoints of the robot.} \label{fig:exp-}
\end{figure}

The robot collects $\bbarm = 16$ initial measurements; these measurements are shown in Figure \ref{fig:exp-meas}. Note that the pattern of readings are in agreement with the model prediction verifying the overall accuracy of the numerical solutions of the flow and the AD-PDE \eqref{eq:BVP}. Figure \ref{fig:exp-place} shows the waypoints where we set the maximum number of measurements to $m_{\max} = 25$.
The final solution is $\bbp = (3140, 1.69, 1.76, 1.77, 1.85)$ resulting in a location error $e_{\text{loc}} = 0.03$. This small error is due to the different heights of the planes at which ethanol is released ($x_3 = 0.30$m) and measurements are taken ($x_3 = 0.27$m). This causes the peak to be somewhat displaced downstream. The intensity of the source is predicted to be approximately $3 \times 10^3$ppm/s for the estimated source area of $5.6 \times 10^{-3}$m$^2$.
The video of the SI process is given in \cite{meJ1_video}. It can be seen from the video that the estimation of the source location approaches the true value immediately after the initial measurements are collected and the rest of the measurements correct the solution for the newer information that becomes available. Particularly, as the measurements get closer to the source, the intensity of the estimated source spikes to account for much larger observed concentration measurements.

\subsection{Discussion} \label{sec:discussion}
An important predicament in application of model-based ASI Algorithm \ref{alg:autoSI} is handling advection-dominated problems. Given a transport medium, e.g., air, high advection translates to turbulent flow which is non-trivial to model and an active area of research. Currently viable approaches are based on the Reynolds-Averaged Navier-Stokes (RANS) models that provide time-averaged properties. These models often suffice for engineering applications but major assumptions used in their derivation, technicalities pertaining to mesh generation and boundary layer treatment, and oftentimes conflicting predictions from different models affect their predictive ability so that additional experimental studies may be necessary to validate them; see \cite{TMCFD1993W} for more details.

On the other hand, advection-dominated AD models are also challenging and an active area of research. The reason is usually the presence of numerical instabilities. 
In the previous section, we added a constant artificial diffusion which is a common practice in the relevant literature \cite{LRSROMBE2017BBSK}. Nevertheless, this might in general lead to forward solutions that are inconsistent with the solution of the original AD-PDE \cite{FEMFP2003DH}.
Numerous more advanced stabilization techniques exist that artificially introduce diffusion in a consistent manner; see e.g. \cite{SUPGFCDF1982BH}. Note that very high Peclet numbers, e.g., $Pe \approx 10^3$, are reported in the literature for the forward solution of the AD-PDE \eqref{eq:BVP} but solving the Inverse Problem using the AD model is considerably more challenging.

Specifically, instability of the AD model adversely affects the POD method, the SA initialization, and consequently the nonlinear optimization problems \eqref{eq:opt} and \eqref{eq:pathOptL}. After extensive simulation and experimental studies, we have observed that our method works well for Peclet numbers up to approximately $Pe \approx 250$.
While here we employ the standard Galerkin scheme for simplicity, more sophisticated FEM could be employed to improve this bound.
Note that an important feature of the ASI Algorithm \ref{alg:autoSI} is that it is highly modular, meaning that different components, i.e., the formulation of the SI problem \eqref{eq:opt}, planning problem \eqref{eq:pathOptL}, model reduction Algorithm \ref{alg:POD}, and SA initialization Algorithm \ref{alg:init}, can be independently improved for better SI performance. For instance in Section \ref{sec:ppComp} we replaced the planning module with the ergodic placement for the purpose of comparison.

Finally, for very low Peclet numbers, i.e., diffusion dominated cases, high concentration regions are local, thus it is possible for the SA method to miss some sources if there are no measurements close enough to those sources. Moreover, in the case of multiple sources, if the intensities differ considerably, the SA technique typically detects the high intensity ones.
In these cases, using higher numbers of initial measurements $\bbarm$ and tuning the thresholding parameter $\alpha$ in the SA Algorithm \ref{alg:init} can improve the initialization. We also note that, the SA initialization can be combined with possible prior knowledge about the sources to improve the performance of the proposed ASI algorithm.

\section{Conclusions} \label{sec:concl}
In this paper we considered the problem of Active Source Identification (ASI) in Advection-Diffusion (AD) transport systems in steady-state.
Specifically, given a set of noisy concentration measurements, we developed a novel feedback control scheme that combines a SI problem and a planning problem to guide a mobile robot through an optimal sequence of measurements allowing it to estimate the desired source.
We employed model reduction and source parametrization techniques to reduce dimensionality and, therefore, the size of the SI and planning problems, and proposed a domain decomposition method to handle non-convex domains and multiple sources.
We illustrated our proposed ASI algorithm in  simulation and real-world experiments.

\section*{Acknowledgements}
The authors would like to thank Eric Stach, Yihui Feng, and Yan Zhang for their help with the design of the experimental setup.

\appendices

\section{Adjoint Method} \label{sec:adjointM}
%
In this section, we discuss the details of the Adjoint Method to obtain the gradient of the SI problem (\ref{eq:opt}) that we formulated in Section \ref{sec:probDef}. Particularly, the Lagrangian function of this constrained optimization problem is given as
$$ \ccalL(c,s,w) = \ccalJ(c,s) + \inprod{w, \ccalM(c,s)} , $$
where $w \in V''$ is the adjoint variable. From reflexivity of the Hilbert space $V$, we get $V'' = V$. Then, referring to the definition of the AD model \eqref{eq:ADmodel}, we have
\begin{eqnarray*}
\inprod{w, \ccalM(c,s)} &=& \inprod{w, Ac - \ell(s)}_{V'' \times V'} \\ 
&=& \inprod{Ac - \ell(s), w}_{V' \times V} = a(c,w) - \ell(w;s) .
\end{eqnarray*}
Thus, we can rewrite the Lagrangian as
\begin{equation} \label{eq:Lag}
\ccalL(c,s,w) = \ccalJ(c,s) + a(c,w) - \ell(w;s) ,
\end{equation}
where $w \in V$ is the adjoint variable.

In what follows, we use the notion of a G\^ateaux derivative to differentiate the Lagrangian (\ref{eq:Lag}); see, e.g.,  \cite[sec. 9.4]{IFA1998R}.
\begin{definition}[G\^ateaux derivative] \label{def:Gateaux}
A functional $\ccalT: V \to \reals$ on a normed space $V$ is G\^ateaux-differentiable at $c \in V$ if there exists an operator $D_c\ccalT: V \to V'$ defined by
$$ \inprod{D_c \ccalT,h} \triangleq \inprod{ \ccalT'_c,h} \triangleq \frac{d}{d \epsilon} [ \ccalT (c+ \epsilon h)] \Big|_{\epsilon = 0} , $$
for all $h \in V$. We use the two notations $\inprod{D_c \ccalT,h}$ and $\inprod{ \ccalT'_c,h}$ interchangeably whenever one of them is clearer.
\end{definition}

The Adjoint Method consists of the following three steps that yield an organized procedure for the calculation of the desired gradient; see, e.g., \cite[sec. 4]{SIPEIUAM2003ANG}.
First, in order to satisfy the AD constraint in the SI problem \eqref{eq:opt}, we set the G\^ateaux derivative of the Lagrangian \eqref{eq:Lag} with respect to the adjoint variable $w$ and in an arbitrary direction $v$ equal to zero. The bilinear form $a(c,w)$ and the functional $\ell(w;s)$ are the terms in the Lagrangian that contain $w$.
G\^ateaux differentiating $a(c,w)$ with respect to $w$ we get
\begin{eqnarray*}
\inprod{D_w a(c,w),v} = \frac{d}{d \epsilon} a(c,w + \epsilon v) \Big|_{\epsilon = 0} = a(c,v) ,
\end{eqnarray*}
where we have used linearity of the bilinear operator $a(c, w)$ in each argument.
Similarly, G\^ateaux differentiating $\ell(w;s)$ with respect to $w$ we get $\inprod{D_w \ell (w;s), v} = \ell(v;s)$.
Therefore, the first equation of the Adjoint Method is given as
\begin{equation} \label{eq:AdM1}
\inprod{\ccalL'_w,v} = a(c,v) - \ell(v;s) = 0, \ \forall v \in V .
\end{equation}
Note that this equation is identical to the VBVP \eqref{eq:VBVP} and for the function $c$ satisfying this equation, i.e., $c = \ccalF(s)$, we get $\ccalL(c,s,w) = \bar{\ccalJ}(s)$. Consequently, we can differentiate the Lagrangian \eqref{eq:Lag} to get the desired derivative $\bar{\ccalJ}'_s$.

Since $c = \ccalF(s)$, in order to calculate $D_s \ccalL(c,s,w)$ we need the derivative $\ccalF'_s$. We can avoid this calculation by setting the G\^ateaux derivative of the Lagrangian \eqref{eq:Lag} with respect to the concentration $c$ equal to zero for any arbitrary direction $h$. The two terms containing $c$ are the objective functional $\ccalJ(c,s)$ and the bilinear form $a(c,w)$.
From Definition \ref{def:Gateaux}, the G\^ateaux derivative of $\ccalJ(c,s)$ with respect to $c$ can be calculated explicitly using equation \eqref{eq:obj} as
\begin{eqnarray} \label{eq:gradJc}
\inprod{\ccalJ'_c, h} = \int_{\Omega} h \, (c - c^m) \ \chi_E \ d\Omega.
\end{eqnarray}
Moreover, similar to the previous case the G\^ateaux derivative of the bilinear form $a(c, w)$  with respect to $c$ is given by
$ \inprod{D_c a(c,w), h}  = a(h,w) = a^*(w,h) , $
where $a^*(w,h)$ is the adjoint operator of the bilinear form $a(h,w)$.
Therefore, the second equation of the Adjoint Method is given as
\begin{equation} \label{eq:AdM2}
\inprod{\ccalL'_c,h} = \inprod{\ccalJ'_c,h} + a^*(w,h) = 0, \ \forall h \in V .
\end{equation}
Because of the appearance of the adjoint operator, this equation is called the adjoint equation and the procedure of calculating the desired gradient is referred to as Adjoint Method. Given the concentration $c$ obtained from \eqref{eq:AdM1}, the solution of equation \eqref{eq:AdM2} yields the corresponding adjoint variable $w$.

From the definition of the Lagrangian \eqref{eq:Lag}, for the functions $c$ and $w$ satisfying equations \eqref{eq:AdM1} and \eqref{eq:AdM2}, we have $D_s \ccalL(c,s,w) = \bar{\ccalJ'_s}$. Thus, we can calculate the desired gradient of the objective functional $\bar{\ccalJ}(s)$ with respect to the source term $s$ in a given direction $q$ by G\^ateaux differentiating the Lagrangian \eqref{eq:Lag} as
%
$ \inprod{\ccalL'_s,q} = \inprod{\ccalJ'_s,q} - \inprod{\ell'_s(w;s),q} .$
%
Combining equations \eqref{eq:AdM1} and \eqref{eq:AdM2} with this equation, we summarize the Adjoint Method to calculate the gradient of $\bar{\ccalJ}(s)$ with respect to $s$ in a given direction $q$ as:
\begin{subequations} \label{eq:adjoint}
\begin{align}
& \text{AD-PDE:} \ \ \ \, a(c,v) - \ell(v;s) = 0 , 						\ \ \ \ \forall v \in V ,  \label{eq:adVBVP} \\
& \text{Adjoint Eq: } \inprod{\ccalJ'_c,h} + a^*(w,h) = 0 , 				\   \forall h \in V,  \label{eq:adAdjoint} \\
& \text{Gradient:} \ \ \ \ \inprod{\ccalL'_s,q} = \inprod{\ccalJ'_s - \ell'_s(w;s),q} . \label{eq:adGrad}
\end{align}
\end{subequations}

\section{Numerical Solution of the\\Source Identification Problem} \label{sec:SInumeric}
\subsection{First Order Information} \label{sec:Grad}
In Appendix \ref{sec:adjointM} we discussed the Adjoint Method to obtain the gradient of the SI problem \eqref{eq:opt} when the variable $s$ is a function that lives in the infinite dimensional function space $S$. Here, we employ the approximations $V_d$ and $S_d$ defined in Section \ref{sec:finiteDim} to obtain a finite dimensional counterpart of the Adjoint Method equations \eqref{eq:adjoint} that is needed to solve the finite dimensional SI problem \eqref{eq:optFD} numerically.

First, we substitute the finite dimensional representations \eqref{eq:finiteDim} into equation \eqref{eq:adVBVP} to get
%
%
\begin{align*}
& a(c_d,v_d) - \inprod{\ell(s_d),v_d} = 0, \ \forall v_d \in V_d, \\
& a( \sum_{k=1}^N c_k \psi_k , \sum_{i=1}^N v_i \psi_i ) - \inprod{\ell(s_d), \sum_{i=1}^N v_i \psi_i } = 0, \ \forall v_i \in \reals, \\
& \sum_{i=1}^N v_i \set{ \sum_{k=1}^N c_k a( \psi_k , \psi_i ) - \inprod{\ell(s_d), \psi_i } } = 0, \ \forall v_i \in \reals, \\
& \sum_{k=1}^N c_k a( \psi_k , \psi_i ) - \inprod{\ell(s_d), \psi_i } = 0 , \ \forall i \in \set{1, \dots, N} .
\end{align*}
Writing the equations for all $i \in \set{1, \dots, N}$ in matrix form, we obtain the following linear system of equations
\begin{equation} \label{eq:VBVPLS}
\bbA \bbc = \bbb (\bbp),
\end{equation}
where $\bbA \in \reals^{N \times N}$ and $\bbb$ is a fixed vector for a given $\bbp$. Using equation \eqref{eq:VBVPLS}, we define the finite dimensional model in equation (\ref{eq:optFD})  explicitly as
\begin{equation} \label{eq:Model}
\bbM(\bbc, \bbp) = \bbA \bbc - \bbb (\bbp) = \bb0 .
\end{equation}
As explained in Section \ref{sec:ADPDE}, the AD model \eqref{eq:VBVP} has a unique solution that translates to the invertibility of matrix $\bbA$ in \eqref{eq:Model}.

Similar to approximations \eqref{eq:finiteDim}, we can write $w_d = \bbpsi \, \bbw \text{ and } \, h_d = \bbpsi \, \bbh$ where $\bbw, \bbh \in \reals^N$. Substituting these definitions into the adjoint equation \eqref{eq:adAdjoint}, we get
\begin{align*}
%
%
& \inprod{\ccalJ'_c, \psi_i } + \sum_{k=1}^N w_k a^*( \psi_k , \psi_i ) = 0 , \ \forall i \in \set{1, \dots, N} ,
\end{align*}
where the derivative $\inprod{\ccalJ'_c, \cdot }$ is defined by equation \eqref{eq:gradJc}.
Again writing the equations for all $i \in \set{1, \dots, N}$ in matrix form, we obtain
\begin{equation} \label{eq:adjointLS}
\bbA^T \bbw = - \bbd,
\end{equation}
where the transpose sign appears in \eqref{eq:adjointLS} because of the adjoint operator in the equations.

Given values for the source parameters $\bbp$, the linear systems \eqref{eq:VBVPLS} and \eqref{eq:adjointLS} can be used to obtain the corresponding concentration $\bbc$ and adjoint variable $\bbw$. This information can then be used in \eqref{eq:adGrad} to calculate the desired gradient $\nabla_{\bbp} \bbarJ$ of the objective function $\bbarJ(\bbp) = \bar{\ccalJ}(s_d)$ with respect to $\bbp$.
In order to simplify the notation and without loss of generality, we assume a single source in a $2$D domain given by
$ s_d(\bbx) = \beta \, \phi(\bbx; \undbbx, \barbx) ,$
where $\undbbx = (\undx_1,\undx_2)$ and $\barbx = (\bbarx_1, \bbarx_2)$.
Substituting the approximations \eqref{eq:finiteDim} in the Lagrangian \eqref{eq:Lag}, we get
$ \ccalL(c_d,s_d,w_d) = \ccalJ(c_d,s_d) + a(c_d,w_d) - \ell(w_d;s_d) .$
To obtain the finite dimensional counterpart of equation \eqref{eq:adGrad}, we need to take the derivative of this Lagrangian with respect to the parameters $\bbp$ of the source term $s_d$. The terms that contain $s_d$ are $\ccalJ(c_d,s_d)$ and $\ell(w_d; s_d)$.
For the objective functional $\ccalJ(c_d,s_d)$ from equation \eqref{eq:obj}, the only part involving $s_d$ is the regularization term
$ \int_{\Omega} s_d \ d\Omega = \beta \ (\bbarx_1 - \undx_1) (\bbarx_2 - \undx_2) .$
From this expression we can calculate the derivatives of $\ccalJ(c_d,s_d)$ with respect to $\bbp$, e.g., 
$$ \frac{\partial \ccalJ}{\partial \undx_1} = - \tau \beta (\bbarx_2 - \undx_2)  . $$
For the functional $\ell(w_d; s_d)$, substituting $s_d$ into the definition \eqref{eq:functional} we get
$ \ell(w_d;s_d) = 
\int_{\undx_1}^{\bbarx_1} \int_{\undx_2}^{\bbarx_2} \beta \ w_d(\bbx) \ dx_2 \ dx_1 . $
The derivative with respect to $\beta$ is straightforward and for the other parameters we use the Leibniz rule, e.g., 
$$ \frac{\partial \ell}{\partial \undx_1} = - \beta \int_{\undx_2}^{\bbarx_2} w_d(\undx_1, x_2) \ dx_2 . $$
Then by equation \eqref{eq:adGrad}, combining the two derivatives for $\undx_1$ we get
$ \partial \bbarJ / \partial \undx_1 =  \partial \ccalJ / \partial \undx_1 -  \partial \ell / \partial \undx_1 .$
The other derivatives can be calculated exactly the same way and we get the following values for the desired gradient
\begin{align} \label{eq:grad}
& \frac{\partial \bbarJ}{\partial \beta} = \displaystyle \tau (\bbarx_1 - \undx_1) (\bbarx_2 - \undx_2) - \int_{\undx_1}^{\bbarx_1} \int_{\undx_2}^{\bbarx_2} w_d(\bbx) \ dx_2 \ dx_1 , \nonumber  \\
& \frac{\partial \bbarJ}{\partial \undx_1} = \displaystyle - \tau \beta (\bbarx_2 - \undx_2) + \beta \int_{\undx_2}^{\bbarx_2} w_d(\undx_1, x_2) \ dx_2 , \nonumber  \\
& \frac{\partial \bbarJ}{\partial \undx_2} = \displaystyle - \tau \beta (\bbarx_1 - \undx_1) + \beta \int_{\undx_1}^{\bbarx_1} w_d(x_1, \undx_2) \ dx_1 , \nonumber  \\
& \frac{\partial \bbarJ}{\partial \bbarx_1} = \displaystyle \tau \beta (\bbarx_2 - \undx_2) - \beta \int_{\undx_2}^{\bbarx_2} w_d(\bbarx_1, x_2) \ dx_2 , \nonumber  \\
& \frac{\partial \bbarJ}{\partial \bbarx_2} = \displaystyle \tau \beta (\bbarx_1 - \undx_1) - \beta \int_{\undx_1}^{\bbarx_1} w_d(x_1, \bbarx_2) \ dx_1 ,
\end{align}
where $\bbarJ(\bbp) = \bar{\ccalJ}(s_d)$ and $\bbp = (\beta, \undx_1, \undx_2, \bbarx_1, \bbarx_2)$.
The process for calculating the desired gradient $\nabla_{\bbp} \bbarJ$ given a set of values for the parameters $\bbp$ is described in Algorithm \ref{alg:Grad}.
\begin{algorithm}[t]
\caption{The Adjoint Method}
\label{alg:Grad}
\begin{algorithmic}[1]

\REQUIRE The vector of parameters $\bbp$ and the matrix $\bbA$;

\STATE Compute the r.h.s. vector $\bbb$ of equation \eqref{eq:VBVPLS};

\STATE Solve the linear system $\bbA \bbc = \bbb$ for coefficients $\bbc$;

\STATE Compute the r.h.s. vector $\bbd$ of equation \eqref{eq:adjointLS} using \eqref{eq:gradJc};

\STATE Solve the adjoint equation $\bbA^T \bbw = - \bbd$ for $\bbw$;

\STATE Compute the desired gradient $\nabla_{\bbp} \bbarJ$ using equation (\ref{eq:grad}).

\end{algorithmic}
\end{algorithm}
Note that if there are multiple sources, i.e., if $M > 1$, then we calculate the gradients for each basis function separately. This follows from the rule for differentiating sums. Moreover if $\Omega \subset \reals^3$, we can exactly follow the same steps to calculate the gradient.

\subsection{Second Order Information} \label{sec:secOrderInfo}
Including second order information in the optimization algorithm can make the solution of the SI problem \eqref{eq:optFD} more efficient and accurate. Such information can be in the form of the Hessian $\bbH = \nabla_{\bbp \bbp} \bbarJ$ of the objective function itself, or in the form of a Hessian-vector product $\bbH  \bbv$, for some vector $\bbv$, that is used in the optimization algorithm; see, e.g., \cite[ch. 7]{NO2006NW}.
The procedure to calculate the Hessian-vector multiplication is an attractive choice for large-scale problems but we use it here since it provides an organized approach to incorporate the AD model \eqref{eq:Model} into the Hessian calculations.
Specifically, using the finite dimensional approximation of the Lagrangian (\ref{eq:Lag}) given as
\begin{equation} \label{eq:LagFiniteDim}
L(\bbc, \bbp, \bbw) = J(\bbc, \bbp) + \bbw^T \bbM(\bbc, \bbp) = J(\bbc, \bbp) + \bbw^T (\bbA \bbc - \bbb) ,
\end{equation}
we can devise a procedure to calculate the product $\bbH  \bbv$ for a given vector $\bbv$.
The details of this derivation can be found in \cite{PAPO2006ABGHK} and it results in the following equation
\begin{equation} \label{eq:HessVec}
\bbH \, \bbv = \bbM_{\bbp}^T \, \bbh_4 + \nabla^2_{\bbp \bbc} L \ \bbh_1 + \nabla^2_{\bbp \bbp} L \ \bbv ,
\end{equation}
where the subscripts denote differentiation and the process to calculate the vectors $\bbh_1$ and $\bbh_4$ is explained in Algorithm \ref{alg:Hess}.
\begin{algorithm}[t]
\caption{Hessian-vector Multiplication}
\label{alg:Hess}
\begin{algorithmic}[1]

\REQUIRE The vector $\bbv$;

\REQUIRE The matrices $\bbA$, $\bbM_{\bbp}$, $\nabla^2_{\bbc \bbc} L$, and $\nabla^2_{\bbp \bbp} L$ from equations \eqref{eq:VBVPLS}, \eqref{eq:Mp}, \eqref{eq:Lcc}, and \eqref{eq:Lpp};

\STATE Compute $\bbh_2 = \bbM_{\bbp} \bbv$ using equation \eqref{eq:Mp};

\STATE Solve $\bbM_{\bbc} \bbh_1 = \bbh_2$ for $\bbh_1$ where $\bbM_{\bbc} = \bbA$;

\STATE Compute $\bbh_3 = \nabla^2_{\bbc \bbp} L \, \bbv + \nabla^2_{\bbc \bbc} L \ \bbh_1 $ using equations \eqref{eq:Lpc} and \eqref{eq:Lcc};

\STATE Solve $\bbM_{\bbc}^{T} \bbh_4 = - \bbh_3$ for $\bbh_4$;

\STATE Calculate $\bbH \, \bbv$ from equation \eqref{eq:HessVec}.

\end{algorithmic}
\end{algorithm}

In what follows, we discuss all the second order terms needed in Algorithm \ref{alg:Hess} starting with the derivative of the AD model \eqref{eq:Model} with respect to the parameters denoted by $\bbM_{\bbp}$.
Recalling equation \eqref{eq:VBVPLS} and using the Leibniz rule, row $i$ of matrix $\bbM_{\bbp} \in \reals^{N \times 5}$ is given as
\begin{align} \label{eq:Mp}
& \frac{\partial \bbM_i}{\partial \beta} = - \int_{\undx_1}^{\bbarx_1} \int_{\undx_2}^{\bbarx_2} \psi_i(\bbx) \ dx_2 \ dx_1 , \nonumber \\
& \frac{\partial \bbM_i}{\partial \undx_1} =   \beta \int_{\undx_2}^{\bbarx_2} \psi_i(\undx_1, x_2) \ dx_2 , \nonumber \\
& \frac{\partial \bbM_i}{\partial \undx_2} =   \beta \int_{\undx_1}^{\bbarx_1} \psi_i(x_1, \undx_2) \ dx_1 , \nonumber \\
& \frac{\partial \bbM_i}{\partial \bbarx_1} = - \beta \int_{\undx_2}^{\bbarx_2} \psi_i(\bbarx_1, x_2) \ dx_2 , \nonumber \\
& \frac{\partial \bbM_i}{\partial \bbarx_2} = - \beta \int_{\undx_1}^{\bbarx_1} \psi_i(x_1, \bbarx_2) \ dx_1 .
\end{align}
Using equation \eqref{eq:Model}, the derivative of the AD model with respect to $\bbc$ is given as
$ \bbM_{\bbc} = \bbA . $
In addition, since there are no terms containing the multiplication of the concentration and source parameters, $\bbc$ and $\bbp$, in $L(\bbc, \bbp, \bbw)$, we have 
\begin{equation} \label{eq:Lpc}
\nabla^2_{\bbp \bbc} L = \nabla^2_{\bbc \bbp} L = \bb0 .
\end{equation}

Finally we need to calculate the second order derivatives of the Lagrangian with respect to each of $\bbc$ and $\bbp$. Note that from equation \eqref{eq:LagFiniteDim}, $\nabla^2_{\bbc \bbc} L = \nabla^2_{\bbc \bbc} J$ and the value $\inprod{\ccalJ'_c,\psi_i}$ is basically the directional derivative in the direction $\psi_i$ or the derivative $\partial J / \partial c_i$. Thus we can Gateaux differentiate equation \eqref{eq:gradJc} once more to get the element in row $i$ and column $j$ of $\nabla^2_{\bbc \bbc} L$ as
\begin{eqnarray} \label{eq:Lcc}
[\nabla^2_{\bbc \bbc} L]_{ij} = \int_{\Omega} \chi_E \, \psi_i \, \psi_j \ d\Omega ,
\end{eqnarray}
where $i, j \in \set{1, \dots, N}$. Note that this expression is independent of the parameters $\bbp$ and can be calculated offline.

In order to calculate $\nabla^2_{\bbp \bbp} L$ note that the terms $J(\bbc, \bbp)$ and $\bbw^T \bbb$ in the Lagrangian (\ref{eq:LagFiniteDim}) contribute to this derivative. The calculation for $\nabla^2_{\bbp \bbp} J$ can be done by differentiating the result of Section \ref{sec:Grad} for $\nabla_{\bbp} J$ once more to get
\begin{equation} \label{eq:Jpp}
\nabla^2_{\bbp \bbp} J = \tau \left[
\begin{array}{ccccc}
0  &  \cdot &  \cdot  &  \cdot  &  \cdot  \\
- (\bbarx_2 - \undx_2)  &  0  & \cdot  &  \cdot  &  \cdot  \\
- (\bbarx_1 - \undx_1)  & \beta  &  0  &  \cdot  &  \cdot  \\
 (\bbarx_2 - \undx_2)  &  0  &  - \beta  &  0   &  \cdot  \\
 (\bbarx_1 - \undx_1)  & - \beta  &  0  &  \beta  &   0
\end{array}
\right] .
\end{equation}
For the second term we have $\bbw^T \bbb = \ell(w_d; s_d) $, since the Lagrangians \eqref{eq:Lag} and \eqref{eq:LagFiniteDim} are equivalent.
Thus, we can differentiate the expression for $\nabla_{\bbp} \ell$ from Section \ref{sec:Grad} once more, using the Leibniz rule, to get $ \nabla^2_{\bbp \bbp} \ell$ as is shown in equation \eqref{eq:LppSec}.
\begin{figure*}[t!]
\small
\begin{align} \label{eq:LppSec}
& \nabla^2_{\bbp \bbp} \ell = \nonumber \\
& \left[
\begin{array}{ccccc}
0  &  \cdot  &  \cdot  &  \cdot  &  \cdot  \\
- \int_{\undx_2}^{\bbarx_2} w_d(\undx_1, x_2) \ dx_2  &  
- \beta \int_{\undx_2}^{\bbarx_2} \frac{\partial w_d}{\partial \undx_1} (\undx_1, x_2) \ dx_2  
&  \cdot  &  \cdot  &  \cdot  \\
- \int_{\undx_1}^{\bbarx_1} w_d(x_1, \undx_2) \ dx_1  & \beta w_d(\undx_1, \undx_2)  &  
- \beta \int_{\undx_1}^{\bbarx_1} \frac{\partial w_d}{\partial \undx_2} (x_1, \undx_2) \ dx_1 
& \cdot  &  \cdot  \\
\int_{\undx_2}^{\bbarx_2} w_d(\bbarx_1, x_2) \ dx_2  &  0  &  - \beta w_d(\bbarx_1, \undx_2)  & 
\beta \int_{\undx_2}^{\bbarx_2} \frac{\partial w_d}{\partial \bbarx_1} (\bbarx_1, x_2) \ dx_2  
&  \cdot  \\
\int_{\undx_1}^{\bbarx_1} w_d(x_1, \bbarx_2) \ dx_1   &  - \beta w_d(\undx_1, \bbarx_2)  &  0  &  \beta w_d(\bbarx_1, \bbarx_2)  &  
\beta \int_{\undx_1}^{\bbarx_1} \frac{\partial w_d}{\partial \bbarx_2} (x_1, \bbarx_2) \ dx_1 
\end{array}
\right]
\end{align}
\end{figure*}
\normalsize
Putting the two terms given by equations \eqref{eq:Jpp} and \eqref{eq:LppSec} together, we have
\begin{equation} \label{eq:Lpp}
\nabla^2_{\bbp \bbp} L = \nabla^2_{\bbp \bbp} J  - \nabla^2_{\bbp \bbp} \ell  .
\end{equation}
Notice that we basically have differentiated equation (\ref{eq:grad}) once more in this process.

The case of multiple sources only affects the terms $\bbM_{\bbp}$ and $\nabla^2_{\bbp \bbp} L$ given by equations (\ref{eq:Mp}) and (\ref{eq:Lpp}), respectively.
Since the source term $s_d$ defined in equation \eqref{eq:sourceFiniteDim} is the summation of nonlinear basis functions, for $\bbM_{\bbp}$ we need to append more columns  using equation \eqref{eq:Mp} corresponding to each basis function. On the other hand, for $\nabla^2_{\bbp \bbp} L$ we have to add blocks of matrices given by equation \eqref{eq:Lpp} corresponding to each basis function to the diagonal of $\nabla^2_{\bbp \bbp} L$.

\subsection{Initialization} \label{sec:init}
Appropriate initialization is critical for the solution of nonlinear optimization problems, such as \eqref{eq:optFD}, since otherwise the solution can get trapped in undesirable local minima.
In this paper, we employ a result on the point-source Sensitivity Analysis (SA) of the SI cost functional, presented in \cite{SSASLSSLS2013SA}, for initialization of our method.
The idea is to determine the sensitivity of the objective functional $\ccalJ(c,s)$ to the appearance of a point source in $\Omega$, i.e., we calculate the derivative of the objective with respect to the point-source term.
The regions with highest sensitivity represent the potential areas where the support of the true source function $\bbars$ is nonzero.
Note that by linearity of the AD-PDE (\ref{eq:BVP}), we only need to consider the infinitesimal deviations of the point-source from zero for a source-free domain, i.e., we calculate the derivative for the constant source function $s=0$.

In \cite{SSASLSSLS2013SA} it is shown that the adjoint variable is a measure of the sensitivity of the cost functional to these infinitesimal changes. Thus given the set of measurements $E$ introduced in Section \ref{sec:probDef}, we can obtain an approximation to the source locations via a solution of the adjoint equation. Specifically, we solve 
$ \bbA^T \barbw = -\barbd $
with
$ \bbard_i = \int_{\Omega}  c^m \ \psi_i \ d\Omega $
for $i \in \set{1, \dots, N}$, to get the desired finite dimensional adjoint function as $\bbarw_d = \bbpsi \, \barbw$.
Then an approximate localization of the source is obtained through thresholding as
\begin{equation} \label{eq:init}
\hhatw_d (\bbx) \triangleq 
\left\{
\begin{array}{ll}
\bbarw_d (\bbx) &  \text{if } \bbarw_d (\bbx) \leq \alpha \, \bbarw^{\min}_d  \\
0			 &  \text{o.w.}
\end{array}
\right. 
\end{equation}
where $\bbarw^{\min}_d = \min_{\bbx \in \Omega} \bbarw_d (\bbx)$ and $\alpha \in (0,1)$.

The thresholding parameter $\alpha$ determines the size of the support of $\hhatw_d (\bbx)$ and thus, the number of compact regions that indicate candidate source locations.
In order to separate these compact regions, we utilize the Single Linkage Agglomerative Clustering (SLAC) algorithm; see, e.g., \cite{MSTSLCA1969GR}. Specifically, given the nodal values $\hhatbbw_d$ of $\hhatw_d (\bbx)$ over the FE-mesh, we cluster the nonzero nodal values into sets $C_k$ for $k \in \set{1, \dots, K}$.
Then, we initialize the SI problem \eqref{eq:optFD} by placing a basis function with a small area at the point with highest sensitivity, given by equation \eqref{eq:init}, in each cluster; see Algorithm \ref{alg:init} for details.
\begin{algorithm}[t]
\caption{Point-source Sensitivity Analysis}
\label{alg:init}
\begin{algorithmic}[1]

\REQUIRE The set of measurements $E$;

\REQUIRE The thresholding parameter $\alpha \in (0,1)$;

\STATE Compute the sensitivity function $\hhatw_d (\bbx)$ from equation \eqref{eq:init} and the set $Z = \set {\bbz_i \, | \, \hhatw_d (\bbz_i) \neq 0, 1 \leq i \leq n}$;

\STATE Divide the set of points $Z$ into $K$ clusters $C_k$ according to their distance using the SLAC algorithm;

\STATE For each cluster $C_k$, set the cluster center as
$$\barbz_k = \argmin_{\bbz_i \in C_k} \hhatw_d (\bbz_i) ;$$

\STATE Initialize the source term \eqref{eq:sourceFiniteDim} using bases $\phi_k(\bbx)$ with small areas centered at $\barbz_k$ and $\beta_k \propto \abs{ \hhatw_d (\barbz_k) }$ .

\end{algorithmic}
\end{algorithm}

\section{Sequential Semi-Definite Programming\\for the Next Best Measurement Problem} \label{sec:SSDP}
\subsection{The Next Best Measurement Problem} \label{sec:NBMP}
In this section we discuss the details of the numerical solution for the path planning developed in Section \ref{sec:path}. Let $\bbF(\bbx) = \bbS^T [ \bbX^T \bbX + \bbpsi(\bbx)^T \bbpsi(\bbx) ] \bbS$ denote the FIM at step $m$, where to simplify notation we have dropped the subscripts. Then introducing an auxiliary variable $z$ we can rewrite the optimization problem \eqref{eq:pathOptL} as
\begin{align} \label{eq:pathOptEig}
& \max_{z, \bbx} \ z \nonumber \\
& \ \st \lambda_i (\bbF(\bbx)) > z, \ \forall  i \in \set{1, \dots, p}, \nonumber \\
& \ \ \ \ \ \ \bbx \in \Omega ,
\end{align}
where $\lambda_i$ denotes the $i$-th eigenvalue of the FIM. Problem \eqref{eq:pathOptEig} can be equivalently written as  
\begin{align} \label{eq:pathOptL2}
& \max_{z, \bbx} \ z 	\nonumber \\
& \ \st \bbF(\bbx) - z \, \bbI \succ \bb0 , 	\nonumber \\
& \ \ \ \ \ \ \bbx \in \Omega ,
\end{align}
where the notation $\succ$ denotes a matrix inequality.
The optimization problem \eqref{eq:pathOptL2} is a nonlinear Semi-Definite Program (SDP) that can be solved using nonlinear optimization techniques; see, e.g., \cite{NO2006R}. In this paper we employ the Sequential SDP (SSDP) method which is the extension of sequential quadratic programming; see, e.g., \cite{SSDPMAPROM2005FJV}.
Defining
\begin{subequations} \label{eq:funDef}
\begin{align}
& f(z, \bbx) \triangleq -z , 	\label{eq:funDeff} \\
& \bbB(z, \bbx) \triangleq (\bbarepsilon+z) \bbI - \bbS^T [ \bbX^T \bbX + \bbpsi(\bbx)^T \bbpsi(\bbx) ] \bbS , \label{eq:funDefB}
\end{align}
\end{subequations}
where $0 < \bbarepsilon \ll 1$ is a very small positive number added to eliminate the strict inequality constraint, we can rewrite problem \eqref{eq:pathOptL2} in standard form as
\begin{align} \label{eq:NLSDP}
& \min_{z, \bbx} \ f(z, \bbx) 	\nonumber \\
& \ \st \bbB(z, \bbx) \preceq \bb0 , 	\nonumber \\
& \ \ \ \ \ \ \bbx \in \Omega .
\end{align}
%
%
The Lagrangian corresponding to problem \eqref{eq:NLSDP} is given as
\begin{equation} \label{eq:pathLag}
L(z, \bbx, \bbLambda) = f(z, \bbx) + ( \bbB(z, \bbx), \bbLambda ) \ ,
\end{equation}
where $\bbLambda \geq \bb0$ is the Lagrange multiplier matrix and the inner-product of two $r \times t$ real matrices $\bbB$ and $\bbLambda$ is defined as 
$$ ( \bbB, \bbLambda ) =  \tr (\bbB^T \bbLambda) = \sum_{i=1}^{r} \sum_{j=1}^{t} b_{ij} \lambda_{ij} . $$
Note that $\bbB(z, \bbx): \reals^{d+1} \to \mbS^p$ in \eqref{eq:funDefB} is a negative-semidefinite symmetric matrix function.
Since the Karush-Kuhn-Tucker (KKT) optimality conditions of the nonlinear SDP \eqref{eq:NLSDP} are locally identical to the second-order approximation around any point $(\bbarz, \barbx, \barbLambda)$, we can solve a sequence of convex SDPs to build the solution of the nonlinear problem \eqref{eq:NLSDP} iteratively. Under certain conditions that are satisfied for the functions in \eqref{eq:funDef}, the SSDP approach converges to a local minimum of the nonlinear SDP \eqref{eq:NLSDP}; see, e.g., \cite{SSDPMAPROM2005FJV}.

Specifically, at each iteration $k$, we construct a second-order convex approximation of \eqref{eq:NLSDP} at point $(z_k, \bbx_k, \bbLambda_k)$ as
\begin{align} \label{eq:tangent}
& \min_{\bbd \in \reals^{d+1}} \ \nabla f(\bbv_k)^T \bbd + 0.5 \ \bbd^T \, \bbH_k \, \bbd  \nonumber \\
& \ \ \ \st \ \ \bbB(\bbv_k) + D_{\bbv} \bbB(\bbv_k) [\bbd] \preceq \bb0 , 	\nonumber \\
& \ \ \ \ \ \ \ \ \ \ \bbx_k + \bbd_{\bbx} \in \Omega ,
\end{align}
where $\bbv_k = (z_k, \bbx_k )$ is the primal variable at iteration $k$ and $\bbd = (d_z, \bbd_{\bbx})$ is a vector of directions, where $d_z \in \reals$ and $\bbd_{\bbx} \in \reals^d$ are the directions corresponding to $z_k$ and $\bbx_k$, respectively. The matrix $\bbH_k$ is a positive semidefinite approximation of the Hessian $ \nabla_{\bbv \bbv}^2 L(z_k, \bbx_k, \bbLambda_k)$ of the Lagrangian \eqref{eq:pathLag} and $D_{\bbv} \bbB(\bbv_k) [\bbd]$ is the directional derivative of the matrix function \eqref{eq:funDefB} at point $\bbv_k$ and direction $\bbd$ that is used to linearize the matrix inequality constraint around the current iterate $\bbv_k$.
This quantity along with the Hessian of the Lagrangian are derived in Appendix \ref{sec:1st2ndDeriv}.
We assume that the domain $\Omega$ is convex and defined by a set of affine constraints so that the linear constraint $\bbx_k + \bbd_{\bbx} \in \Omega$ can be directly incorporated in the SDP \eqref{eq:tangent}. This assumption holds for the box constrained domain that we considered in the SI problem \eqref{eq:optFD}.

The solution of the SDP \eqref{eq:tangent} denoted by $\bbd_k \in \reals^{d+1}$ determines the descent direction for the nonlinear problem \eqref{eq:NLSDP}. Using this solution, we update the primal variables as
\begin{equation} \label{eq:iterSSDP}
\bbv_{k+1} = \bbv_k + \alpha_k \bbd_k ,
\end{equation}
where $\alpha_k$ is a step-size whose selection is explained in Appendix \ref{sec:stepSize}.
Note that by the last constraint in the SDP problem \eqref{eq:tangent}, we implicitly assume that the maximum step-size is equal to one, i.e., $\alpha_{\max}=1$.
We update the dual variable $ \bbLambda_{k+1}$ directly as the optimal dual of the tangent problem \eqref{eq:tangent}. The details of the SSDP to solve the optimization problem \eqref{eq:pathOptL} are presented next.

\subsection{First and Second Order Information} \label{sec:1st2ndDeriv}
To solve problem \eqref{eq:NLSDP} we need the gradient and Hessian information. For this, we first define the expressions $D_{\bbv} \bbB(\bbv) [\bbd]$ and $\nabla_{\bbv \bbv}^2 L(z, \bbx, \bbLambda)$ that appear in the convex second-order SDP \eqref{eq:tangent}.
For the first term, we have
\begin{equation} \label{eq:linearCons}
D_{\bbv} \bbB(\bbv) [\bbd] = \sum_{i=1}^{d+1} d_i \ \bbB^{(i)}(\bbv) \, , 
\end{equation}
where $d_i$ is the $i$-th element of the vector of directions $\bbd$ and
\begin{equation} \label{eq:deriv1}
\bbB^{(i)}(\bbv) = \frac{\partial}{\partial v_i} \bbB(\bbv) ,
\end{equation}
for $i \in \set{1, \dots, d+1}$.
The operator $D_{\bbv} \bbB(\bbv): \reals^{d+1} \to \mbS^p$ is linear in $\bbd$ and $ D_{\bbv} \bbB(\bbv) [\bbd] \in \mbS^p$. Therefore the corresponding constraint in the SDP \eqref{eq:tangent} is a linear matrix inequality.

For the second term, i.e., the Hessian $\nabla_{\bbv \bbv}^2 L(z, \bbx, \bbLambda)$ of the Lagrangian \eqref{eq:pathLag}, since the objective function $f(z, \bbx)$ defined by equation \eqref{eq:funDeff} is linear, we have
$$\nabla_{\bbv \bbv}^2 L(z, \bbx, \bbLambda) = \nabla_{\bbv \bbv}^2 ( \bbB(z, \bbx), \bbLambda ) \in \mbS^{d+1} ,$$
where
\begin{align} \label{eq:LagHess}
\nabla_{\bbv \bbv}^2 ( \bbB, \bbLambda ) =
\left[
\begin{array}{ccc}
( \bbB^{(1,1)}, \bbLambda )  & \dots  & ( \bbB^{(1,d+1)}, \bbLambda )  \\
\vdots  &  \ddots & \vdots  \\
( \bbB^{(d+1,1)}, \bbLambda )  & \dots  & ( \bbB^{(d+1,d+1)}, \bbLambda )
\end{array}
\right] ,
 \end{align}
and
\begin{equation} \label{eq:deriv2}
\bbB^{(i,j)}(\bbv) = \frac{\partial^2}{\partial v_i \partial v_j} \bbB(\bbv) . 
\end{equation}

Recalling the definition of the matrix function  $\bbB(z, \bbx)$, given in equation \eqref{eq:funDefB}, we calculate the required derivatives \eqref{eq:deriv1} and \eqref{eq:deriv2} for, e.g., the $2$D case in which $\bbx = (x_1, x_2)$.
These derivatives then are used in  equation \eqref{eq:tangent} to build quadratic SDPs that we solve sequentially to find the local optimum of the nonlinear SDP \eqref{eq:NLSDP}.

For the first order derivatives used in equation \eqref{eq:linearCons}, we have
\begin{align*}
& \bbB^{(1)} =  \frac{\partial \bbB}{\partial z} = \bbI, \\
& \bbB^{(2)} =  \frac{\partial \bbB}{\partial x_1} = - \bbS^T \left[ (\frac{\partial \bbpsi}{\partial x_1})^T \bbpsi + \bbpsi^T \frac{\partial \bbpsi}{\partial x_1} \right] \bbS .
\end{align*}
The value for $\bbB^{(3)}$ is exactly the same as $\bbB^{(2)}$, except that the differentiation variable is $x_2$.
Similarly for the second-order derivatives used in equation \eqref{eq:LagHess}, we have
$$ \bbB^{(1,1)} = \bbB^{(2,1)} = \bbB^{(3,1)} = \bb0 , $$
$$  \bbB^{(2,2)} = - \bbS^T \left\{ (\frac{\partial^2 \bbpsi}{\partial x_1^2})^T \bbpsi + 2 (\frac{\partial \bbpsi}{\partial x_1})^T \frac{\partial \bbpsi}{\partial x_1} + \bbpsi^T \frac{\partial^2 \bbpsi}{\partial x_1^2} \right\} \bbS. $$
$\bbB^{(3,3)}$ can be calculated exactly the same way. Finally, for the cross-derivative we have
\begin{align*}
\bbB^{(3,2)} = - \bbS^T & \left\{ (\frac{\partial^2 \bbpsi}{\partial x_1 x_2})^T \bbpsi + (\frac{\partial \bbpsi}{\partial x_1})^T \frac{\partial \bbpsi}{\partial x_2} + \right. \\
& \left. (\frac{\partial \bbpsi}{\partial x_2})^T \frac{\partial \bbpsi}{\partial x_1} + \bbpsi^T \frac{\partial^2 \bbpsi}{\partial x_1 x_2}  \right\} \bbS.
\end{align*}

After calculating the Hessian $\nabla_{\bbv \bbv}^2 L(z, \bbx, \bbLambda)$ of the Lagrangian \eqref{eq:pathLag}, we construct a positive-definite approximation $\bbH$ of it so that the SDP \eqref{eq:tangent} is strictly convex with a unique global minimizer $\bbd$.
Such an approximation of $\bbH$ can be obtained in different ways; see, e.g., \cite{MIMAO1998HC}.
Here, we add a multiple of the identity matrix so that the minimum eigenvalue is bounded from zero by a small amount $\delta$, i.e., we set
\begin{equation} \label{eq:convexification}
\bbH =  \nabla_{\bbv \bbv}^2 L + \mu \, \bbI,
\end{equation}
where $\mu = \max(0, \delta - \lambda_{\min}(\nabla_{\bbv \bbv}^2 L))$. The positive-definite matrix $\bbH$ is the closest to the Hessian $\nabla_{\bbv \bbv}^2 L$ measured by the induced Euclidean norm. Note that since the Hessian is a low dimensional matrix, i.e., $\nabla_{\bbv \bbv}^2 L \in \mbS^{d+1}$, we can easily calculate its minimum eigenvalue.

\subsection{Step-Size Selection} \label{sec:stepSize}
Necessary for the solution of the nonlinear SDP \eqref{eq:NLSDP} is an effective line-search strategy that connects the successive solutions of the quadratic SDPs \eqref{eq:tangent}. In this paper, we utilize the results from \cite{GANSDP2004C} to select an appropriate step-size $\alpha_k$ for the iterations of the SSDP defined by equation \eqref{eq:iterSSDP}.
The final SSDP algorithm to solve the nonlinear SDP \eqref{eq:NLSDP} is presented in Algorithm \ref{alg:SSDP}.
\begin{algorithm}[t]
\caption{Sequential Semi-definite Programming}
\label{alg:SSDP}
\begin{algorithmic}[1]

\REQUIRE The POD bases $\bbpsi = [\psi_1, \dots, \psi_N]$ of Algorithm \ref{alg:POD};

\REQUIRE The parameters $\bbarepsilon$ and $\delta$ of equations \eqref{eq:funDef} and \eqref{eq:convexification};

\REQUIRE The parameters $\epsilon_1$, $\epsilon_2$,  and $\epsilon_3$ of equation \eqref{eq:stop};

\REQUIRE The parameters $\bbargamma > 0$, $\rho \in (0,1)$, and $\omega \in (0,1)$;

\STATE Initialize the iteration index $k=0$;

\STATE Initialize the primal variable $\bbv_0$ using equation \eqref{eq:initialize} and the dual variable $\bbLambda_0$ with $\bbLambda_0 \succeq \bb0$; 		\label{line:initialize}

\STATE Initialize the penalty parameter as $\gamma_0 = \tr(\bbLambda_0) + \bbargamma$;

\WHILE{the algorithm has not converged}

\STATE Build the convex SDP \eqref{eq:tangent} at $(\bbv_k, \bbLambda_k)$ using equations \eqref{eq:linearCons}, \eqref{eq:LagHess}, and \eqref{eq:convexification} and solve it for $(\bbd_k, \bbLambda_{k+1})$;

\STATE Check the stopping criterion \eqref{eq:stop} for $\bbv_k$;

\STATE Set $\gamma_k = \gamma_{k-1}$ if $\set{ \gamma_{k-1} \geq \tr(\bbLambda_{k+1}) + \bbargamma }$, otherwise set it as $\gamma_k = \max \set{1.5 \gamma_{k-1}, \tr(\bbLambda_{k+1}) + \bbargamma}$;

\STATE Select $\alpha_k$ as the largest member of the geometric sequence $\set{1, \rho, \rho^2, \dots}$ such that 
$$ \theta_{\gamma_k} (\bbv_k + \alpha_k \bbd_k) \leq \theta_{\gamma_k} (\bbv_k) + \omega \alpha_k \Delta_k , $$
where the penalty function $\theta_{\gamma}(\bbv)$ and $\Delta_k$ are defined in equations \eqref{eq:penFun} and \eqref{eq:upperB}, respectively;			\label{line:Armijo}

\STATE Update the primal variable $\bbv_{k+1}$ by equation \eqref{eq:iterSSDP};

\STATE $k \leftarrow k+1$;

\ENDWHILE

\end{algorithmic}
\end{algorithm}
In this algorithm, we define the penalty function $\theta_{\gamma}(\bbv)$ for the selection of the step-size $\alpha_k$ as
\begin{equation} \label{eq:penFun}
\theta_{\gamma}(\bbv) = f(\bbv) + \gamma \, \lambda_{\max}(\bbB(\bbv))_+ ;
\end{equation}
where $\gamma > 0$ is the penalty parameter, $\lambda_{\max}(\bbB)_+ = \max \set{0, \lambda_{\max}(\bbB)}$, and the functions $f(\bbv)$ and $\bbB(\bbv)$ are defined in equation \eqref{eq:funDef}. The upper bound $\Delta_k$ on the directional derivative $\theta'_{\gamma_k}(\bbv_k; \bbd_k)$ of the penalty function in a direction $\bbd_k$ is given as 
\begin{equation} \label{eq:upperB}
\Delta_k = - \bbd_k^T \bbH_k \, \bbd_k + \tr(\bbLambda_{k+1} \bbB(\bbv_k)) - \gamma_k \, \lambda_{\max}(\bbB(\bbv_k))_+ ,
\end{equation}
where $\bbH_k$ is the positive-definite approximation given by equation \eqref{eq:convexification} and we have included the index $k$ to emphasize that we use the dual variable $\bbLambda_{k+1}$ to select the step-size $\alpha_k$.
The upper bound $\Delta_k$ is used in order to satisfy the Armijo condition in the backtracking line-search corresponding to line \ref{line:Armijo} in Algorithm \ref{alg:SSDP}. For theoretical details see \cite{GANSDP2004C}.

Note that since the domain of interest $\Omega$ is represented by a set of affine constraints that require no further linearization, the constraint $\bbx \in \Omega$ does not appear in the penalty function \eqref{eq:penFun}. Essentially, the constraint $\bbx \in \Omega$ is never violated and thus we do not penalize it in \eqref{eq:penFun}.

\subsection{Initialization and Stopping} \label{sec:pathInit}
Since the eigenvalue optimization problem \eqref{eq:pathOptL} is nonlinear, appropriate initialization of Algorithm \ref{alg:SSDP} is critical to obtain a reasonable solution. 
Moreover, addition of a new measurement, reshapes the objective function \eqref{eq:pathOptL} and makes it flat around that measurement location. In other words, adding more measurements in that vicinity does not provide more information about the unknown source parameters compared to farther locations.
Therefore without global knowledge of the objective function, the algorithm gets trapped in undesirable local minima where the objective function does not change no matter how many measurements are taken in that region.

In order to generate new informative measurements, we sample the objective function of the Next Best Measurement Problem \eqref{eq:pathOptL}, 
$ g(\bbx) = \lambda_{\min} [ \bbF(\bbp) + \bbS(\bbp)^T \bbpsi(\bbx)^T \bbpsi(\bbx) \bbS(\bbp) ] , $ 
 over a coarse set of points $\bbz_i \in \reals^d$ from the FE mesh, where $i \in \set{1, \dots, Z}$ for some $Z \ll n$ and $n$ is the number of FE grid points.
Then to initialize the primal variable $\bbv_0 = (z_0, \bbx_0)$ in Algorithm \ref{alg:SSDP} for step $m + 1$ of the robot, we calculate the values of the objective function $g_i = g(\bbz_i)$ over this set of points and we set
\begin{equation} \label{eq:initialize}
z_0 = \max_i g_i \ \text{and} \ \bbx_0 = \argmax_{\bbz_i} g(\bbz_i) ,
\end{equation}
where $z_0$ is the auxiliary variable introduced in \eqref{eq:pathOptEig}.
%
Note that each evaluation of the function $g(\bbx)$ amounts to solving a minimum eigenvalue problem for a $p \times p$ matrix where $p$ is the number of unknown parameters. The computational cost of such sampling procedure is comparable to a single backtracking line-search step of Algorithm \ref{alg:SSDP} in line \ref{line:Armijo}.


Finally, to determine whether Algorithm \ref{alg:SSDP} has reached a local minimum we evaluate bounds on the gradient of Lagrangian \eqref{eq:pathLag}, the nonlinear matrix inequality constraint violation, and the complementarity condition as follows
\begin{align} \label{eq:stop}
& \norm{ \nabla_{\bbv} L(\bbv_k, \bbLambda_{k+1}) }_2 \leq \epsilon_1, \ \lambda_{\max}(\bbB(\bbv_k))_+ \leq \epsilon_2, \nonumber \\
& \ \ \ \ \ \ \ \ \ \ \ \ \ \ \ \abs{ \tr(\bbLambda_{k+1} \bbB(\bbv_k)) } \leq \epsilon_3 .
\end{align}
See \cite{GANSDP2004C} for theoretical results supporting this selection of stopping criteria.

\ifCLASSOPTIONcaptionsoff
  \newpage
\fi

\bibliographystyle{ieeetr}
\bibliography{MyBibliography}

\begin{thebibliography}{10}

\bibitem{ROL2008KR}
G.~Kowadlo and R.~A. Russell, ``Robot odor localization: a taxonomy and
  survey,'' {\em International Journal of Robotics Research}, vol.~27, no.~8,
  pp.~869--894, 2008.

\bibitem{OMRN2002MND}
L.~Marques, U.~Nunes, and A.~T. de~Almeida, ``Olfaction-based mobile robot
  navigation,'' {\em Thin solid films}, vol.~418, no.~1, pp.~51--58, 2002.

\bibitem{CRRCA2003RBSW}
R.~A. Russell, A.~Bab-Hadiashar, R.~L. Shepherd, and G.~G. Wallace, ``A
  comparison of reactive robot chemotaxis algorithms,'' {\em Robotics and
  Autonomous Systems}, vol.~45, no.~2, pp.~83--97, 2003.

\bibitem{BAASGTCP2012WVW}
D.~Webster, K.~Volyanskyy, and M.~Weissburg, ``Bioinspired algorithm for
  autonomous sensor-driven guidance in turbulent chemical plumes,'' {\em
  Bioinspiration \& biomimetics}, vol.~7, no.~3, p.~036023, 2012.

\bibitem{BREM2004DSR}
A.~Dhariwal, G.~S. Sukhatme, and A.~A. Requicha, ``Bacterium-inspired robots
  for environmental monitoring,'' in {\em Proceedings of {IEEE} International
  Conference on Robotics and Automation}, vol.~2, pp.~1436--1443, IEEE, 2004.

\bibitem{DRACPT2005ZSS}
D.~Zarzhitsky, D.~F. Spears, and W.~M. Spears, ``Distributed robotics approach
  to chemical plume tracing,'' in {\em Proceedings of {IEEE} International
  Conference on Intelligent Robots and Systems}, pp.~4034--4039, 2005.

\bibitem{DFOSLA2015SAPM}
J.~M. Soares, A.~P. Aguiar, A.~M. Pascoal, and A.~Martinoli, ``A distributed
  formation-based odor source localization algorithm-design, implementation,
  and wind tunnel evaluation,'' in {\em Proceedings of {IEEE} International
  Conference on Robotics and Automation}, pp.~1830--1836, IEEE, 2015.

\bibitem{VCPMOSL2013CMG}
G.~Cabrita, L.~Marques, and V.~Gazi, ``Virtual cancelation plume for multiple
  odor source localization,'' in {\em Proceedings of {IEEE} International
  Conference on Intelligent Robots and Systems}, pp.~5552--5558, IEEE, 2013.

\bibitem{infotaxis2007VVS}
M.~Vergassola, E.~Villermaux, and B.~I. Shraiman, ```{I}nfotaxis' as a strategy
  for searching without gradients,'' {\em Nature}, vol.~445, no.~7126, p.~406,
  2007.

\bibitem{MSSLTM2016HHS}
H.~Hajieghrary, M.~A. Hsieh, and I.~B. Schwartz, ``Multi-agent search for
  source localization in a turbulent medium,'' {\em Physics Letters A},
  vol.~380, no.~20, pp.~1698--1705, 2016.

\bibitem{DIP2010H}
P.~C. Hansen, {\em Discrete Inverse Problems}.
\newblock {SIAM}, 2010.

\bibitem{SLSDENAD2005MGK}
J.~Matthes, L.~Groll, and H.~B. Keller, ``Source localization by spatially
  distributed electronic noses for {A}dvection and {D}iffusion,'' {\em {IEEE}
  Transactions on Signal Processing}, vol.~53, pp.~1711--1719, May 2005.

\bibitem{MBSTSLP2000AS}
M.~Alpay and M.~Shor, ``Model-based solution techniques for the source
  localization problem,'' {\em {IEEE} Transactions on Control Systems
  Technology}, vol.~8, pp.~895--904, Nov 2000.

\bibitem{MSDLADPUWSN2009WSK}
J.~Weimer, B.~Sinopoli, and B.~H. Krogh, ``Multiple source detection and
  localization in {A}dvection-{D}iffusion processes using wireless sensor
  networks,'' in {\em {IEEE} Real-Time Systems Symposium}, pp.~333--342, Dec
  2009.

\bibitem{SEUIRRHSMO2012Z}
F.~Zhu, ``State estimation and unknown input reconstruction via both
  reduced-order and high-order sliding mode observers,'' {\em Journal of
  Process Control}, vol.~22, no.~1, pp.~296--302, 2012.

\bibitem{meC4}
R.~Khodayi-mehr, W.~Aquino, and M.~M. Zavlanos, ``Distributed reduced order
  source identification,'' in {\em Proceedings of American Control Conference},
  pp.~1084--1089, June 2018.

\bibitem{VFEMSICDT2003ABGLW}
V.~Ak{\c{c}}elik, G.~Biros, O.~Ghattas, K.~R. Long, and B.~van
  Bloemen~Waanders, ``A variational finite element method for source inversion
  for {C}onvective-{d}iffusive transport,'' {\em Finite Elements in Analysis
  and Design}, vol.~39, pp.~683--705, May 2003.

\bibitem{meC1}
R.~Khodayi-mehr, W.~Aquino, and M.~M. Zavlanos, ``Model-based sparse source
  identification,'' in {\em Proceedings of American Control Conference},
  pp.~1818--1823, July 2015.

\bibitem{meJ2}
R.~Khodayi-mehr, Y.~Kantaros, and M.~M. Zavlanos, ``Distributed state
  estimation using intermittently connected robot networks,'' {\em {IEEE}
  Transactions on Robotics}, 2018.
\newblock (accepted). [Online]. Available:
  https://arxiv.org/pdf/1805.01574.pdf.

\bibitem{EEDI2016MSMM}
L.~M. Miller, Y.~Silverman, M.~A. MacIver, and T.~D. Murphey, ``Ergodic
  exploration of distributed information,'' {\em {IEEE} Transactions on
  Robotics}, vol.~32, no.~1, pp.~36--52, 2016.

\bibitem{PFBIAS2010RH}
A.~Ryan and J.~K. Hedrick, ``Particle filter based information-theoretic active
  sensing,'' {\em Robotics and Autonomous Systems}, vol.~58, no.~5,
  pp.~574--584, 2010.

\bibitem{MIPPASEP2014CLD}
N.~Cao, K.~H. Low, and J.~M. Dolan, ``Multi-robot informative path planning for
  active sensing of environmental phenomena: A tale of two algorithms,'' in
  {\em International Conference on Autonomous Agents and Multi-agent Systems},
  pp.~7--14, 2013.

\bibitem{DASSSMRN2015ALP}
N.~A. Atanasov, J.~L. Ny, and G.~J. Pappas, ``Distributed algorithms for
  stochastic source seeking with mobile robot networks,'' {\em Dynamic Systems,
  Measurement, and Control}, vol.~137, no.~3, p.~031004, 2015.

\bibitem{SMPFT2011AL}
J.~M. Aughenbaugh and B.~R. LaCour, ``Sensor management for particle filter
  tracking,'' {\em {IEEE} Transactions on Aerospace and Electronic Systems},
  vol.~47, no.~1, pp.~503--523, 2011.

\bibitem{SRM2010FMWB}
M.~Frangos, Y.~Marzouk, K.~Willcox, and B.~van Bloemen~Waanders, ``Surrogate
  and reduced-order modeling: A comparison of approaches for large-scale
  statistical inverse problems,'' {\em Large Scale Inverse Problems and
  Quantification of Uncertainty}, vol.~123149, 2010.

\bibitem{OMMDPSI2004D}
D.~Ucinski, {\em Optimal measurement methods for distributed parameter system
  identification}.
\newblock CRC Press, 2004.

\bibitem{ASIGRPE2005CR}
V.~Christopoulos and S.~Roumeliotis, ``Adaptive sensing for instantaneous gas
  release parameter estimation,'' in {\em Proceedings of {IEEE} International
  Conference on Robotics and Automation}, pp.~4450--4456, April 2005.

\bibitem{OSMSNDLSCS2008PU}
M.~Patan and D.~Uciriski, ``Optimal scheduling of mobile sensor networks for
  detection and localization of stationary contamination sources,'' in {\em
  International Conference on Information Fusion}, pp.~1--7, June 2008.

\bibitem{OMSPPEDPS2008TC}
C.~Tricaud and Y.~Chen, ``Optimal mobile sensing policy for parameter
  estimation of distributed parameter systems: finite horizon closed-loop
  solution,'' in {\em Proceedings of International Symposium on Mathematical
  Theory of Networks and Systems}, 2008.

\bibitem{DPSUE2009BLTT}
M.~Burger, Y.~Landa, N.~M. Tanushev, and R.~Tsai, ``Discovering a point source
  in unknown environments,'' in {\em Algorithmic Foundation of Robotics
  {VIII}}, pp.~663--678, Springer, 2009.

\bibitem{OFROMPOD2007A}
W.~Aquino, ``An object-oriented framework for reduced-order models using
  {P}roper {O}rthogonal {D}ecomposition ({POD}),'' {\em Computer Methods in
  Applied Mechanics and Engineering}, vol.~196, no.~41, pp.~4375--4390, 2007.

\bibitem{SSASLSSLS2013SA}
A.~Sabelli and W.~Aquino, ``A source sensitivity approach for source
  localization in steady-state linear systems,'' {\em Inverse Problems},
  vol.~29, no.~9, p.~095005, 2013.

\bibitem{TMCFD1993W}
D.~C. Wilcox, {\em Turbulence modeling for CFD}, vol.~2.
\newblock DCW Industries La Canada, CA, 1993.

\bibitem{TD2002RW}
P.~J. Roberts and D.~R. Webster, {\em Turbulent diffusion}.
\newblock ASCE Press, Reston, Virginia, 2002.

\bibitem{meC2}
R.~Khodayi-mehr, W.~Aquino, and M.~M. Zavlanos, ``Nonlinear reduced order
  source identification,'' in {\em Proceedings of American Control Conference},
  pp.~6302--6307, July 2016.

\bibitem{STMTPE2005SJ}
S.~A. Socolofsky and G.~H. Jirka, {\em Special topics in mixing and transport
  processes in the environment}.
\newblock Coastal and Ocean Engineering Division, Texas A\&M University, 5~ed.,
  2005.

\bibitem{IFA1998R}
B.~D. Reddy, {\em Introductory functional analysis: with applications to
  boundary value problems and finite elements}, vol.~27.
\newblock Springer, 2013.

\bibitem{ALDS2005A}
A.~C. Antoulas, {\em Approximation of large-scale dynamical systems}, vol.~6.
\newblock {SIAM}, 2005.

\bibitem{PODRBFCPE2001AK}
J.~A. Atwell and B.~B. King, ``Proper {O}rthogonal {D}ecomposition for reduced
  basis feedback controllers for parabolic equations,'' {\em Mathematical and
  Computer Modeling}, vol.~33, no.~1, pp.~1--19, 2001.

\bibitem{NSPMIMCC2006KGGK}
A.~Krause, C.~Guestrin, A.~Gupta, and J.~Kleinberg, ``Near-optimal sensor
  placements: Maximizing information while minimizing communication cost,'' in
  {\em International Conference on Information Processing in Sensor Networks},
  pp.~2--10, ACM, 2006.

\bibitem{CVX}
M.~Grant and S.~Boyd, ``{CVX}: Matlab software for disciplined convex
  programming, version 2.0.'' \url{http://cvxr.com/cvx}, Mar. 2014.

\bibitem{NO2006NW}
J.~Nocedal and S.~J. Wright, {\em Numerical Optimization}.
\newblock New York: Springer-Verlag, 2006.

\bibitem{FCFEM2007L}
D.~L. Logan, {\em A First Course in the Finite Element Method}.
\newblock Nelson, 2007.

\bibitem{cubit}
B.~Hanks, ``{CUBIT} toolkit.'' \url{https://cubit.sandia.gov}.

\bibitem{CPDENMDP2003L}
H.~Langtangen, {\em Computational Partial Differential Equations - Numerical
  Methods and Diffpack Programming}.
\newblock Springer, 2~ed., 2003.

\bibitem{MEDEDMAS2011MM}
G.~Mathew and I.~Mezi{\'c}, ``Metrics for ergodicity and design of ergodic
  dynamics for multi-agent systems,'' {\em Physica D: Nonlinear Phenomena},
  vol.~240, no.~4, pp.~432--442, 2011.

\bibitem{VisiLibity2008O}
K.~J. Obermeyer and Contributors, ``{VisiLibity}: A {C}++ library for
  visibility computations in planar polygonal environments.''
  http://www.VisiLibity.org, 2008.
\newblock R-1.

\bibitem{meJ1_video}
``Video clip: \url{https://vimeo.com/294258707}.''

\bibitem{LRSROMBE2017BBSK}
M.~Benosman, J.~Borggaard, O.~San, and B.~Kramer, ``Learning-based robust
  stabilization for reduced-order models of 2d and 3d boussinesq equations,''
  {\em Applied Mathematical Modelling}, vol.~49, pp.~162--181, 2017.

\bibitem{FEMFP2003DH}
J.~Donea and A.~Huerta, {\em Finite element methods for flow problems}.
\newblock John Wiley \& Sons, 2003.

\bibitem{SUPGFCDF1982BH}
A.~N. Brooks and T.~J. Hughes, ``Streamline upwind/{P}etrov-{G}alerkin
  formulations for convection dominated flows with particular emphasis on the
  incompressible navier-stokes equations,'' {\em Computer methods in applied
  mechanics and engineering}, vol.~32, no.~1-3, pp.~199--259, 1982.

\bibitem{SIPEIUAM2003ANG}
A.~A. Oberai, N.~H. Gokhale, and G.~R. Feijoo, ``Solution of inverse problems
  in elasticity imaging using the {A}djoint method,'' {\em Inverse Problems},
  vol.~19, no.~2, p.~297, 2003.

\bibitem{PAPO2006ABGHK}
V.~Ak{\c{c}}elik, G.~Biros, O.~Ghattas, J.~Hill, D.~Keyes, and B.~van
  Bloemen~Waanders, ``Parallel algorithms for {PDE}-constrained optimization,''
  {\em {SIAM} Parallel Processing for Scientific Computing}, vol.~20,
  pp.~291--320, 2006.

\bibitem{MSTSLCA1969GR}
J.~C. Gower and G.~Ross, ``Minimum spanning trees and single linkage cluster
  analysis,'' {\em Applied statistics}, pp.~54--64, 1969.

\bibitem{NO2006R}
A.~P. Ruszczy{\'n}ski, {\em Nonlinear optimization}, vol.~13.
\newblock Princeton University Press, 2006.

\bibitem{SSDPMAPROM2005FJV}
R.~W. Freund, F.~Jarre, and C.~Vogelbusch, ``A sequential semidefinite
  programming method and an application in passive reduced-order modeling,''
  {\em arXiv Preprint Math}, no.~0503135, 2005.

\bibitem{MIMAO1998HC}
N.~J. Higham and S.~H. Cheng, ``Modifying the inertia of matrices arising in
  optimization,'' {\em Linear Algebra and its Applications}, vol.~275,
  pp.~261--279, 1998.

\bibitem{GANSDP2004C}
R.~Correa, ``A global algorithm for nonlinear semidefinite programming,'' {\em
  {SIAM} Journal on Optimization}, vol.~15, no.~1, pp.~303--318, 2004.

\end{thebibliography}

\end{document}